\documentclass[acmsmall,screen]{acmart}

\AtBeginDocument{%
  }

\setcopyright{acmcopyright}
\copyrightyear{2023}
\acmYear{2023}
\acmDOI{XXXXXXX.XXXXXXX}

\acmJournal{JACM}
\acmVolume{37}
\acmNumber{4}
\acmArticle{111}
\acmMonth{12}

\usepackage{algorithm}
\usepackage{algorithmic}
\usepackage{graphicx}
\usepackage{amsmath}
\usepackage{bm}
\usepackage{multirow}
\usepackage{enumitem}
\usepackage[switch]{lineno} 
\usepackage{animate}

\begin{document}

\title{Exploiting Spatial-Temporal Context for Interacting Hand Reconstruction on Monocular RGB Video}

\author{Weichao Zhao}
\email{saruka@mail.ustc.edu.cn}
\affiliation{%
  \institution{University of Science and Technology of China}
  \city{Hefei}
  \state{AnHui}
  \country{China}
  \postcode{230027}
}

\author{Hezhen Hu}
\email{alexhu@mail.ustc.edu.cn}
\affiliation{%
	\institution{University of Science and Technology of China}
	\city{Hefei}
	\state{AnHui}
	\country{China}
}

\author{Wengang Zhou}
\email{zhwg@ustc.edu.cn}
\affiliation{%
	\institution{University of Science and Technology of China}
	\city{Hefei}
	\state{AnHui}
	\country{China}
}

\author{Li Li}
\email{lil1@ustc.edu.cn}
\affiliation{%
	\institution{University of Science and Technology of China}
	\city{Hefei}
	\state{AnHui}
	\country{China}
}

\author{Houqiang Li}
\email{lihq@ustc.edu.cn}
\affiliation{%
	\institution{University of Science and Technology of China}
	\city{Hefei}
	\state{AnHui}
	\country{China}
}

\renewcommand{\shortauthors}{Weichao Zhao et al.}

\begin{abstract}
  Reconstructing interacting hands from monocular RGB data is a challenging task, as it involves many interfering factors, \emph{e.g.} self- and mutual occlusion and similar textures. 
  Previous works only leverage information from a single RGB image without modeling their physically plausible relation, which leads to inferior reconstruction results.
  In this work, we are dedicated to explicitly exploiting spatial-temporal information to achieve better interacting hand reconstruction.
  On one hand, we leverage temporal context to complement insufficient information provided by the single frame, and design a novel temporal framework with a temporal constraint for interacting hand motion smoothness.
  On the other hand, we further propose an interpenetration detection module to produce kinetically plausible interacting hands without physical collisions.
  Extensive experiments are performed to validate the effectiveness of our proposed framework, which achieves new state-of-the-art performance on public benchmarks.
\end{abstract}

\ccsdesc[500]{Computing methodologies}
\ccsdesc[500]{Artificial intelligence~Computer vision}

\keywords{interacting hand, model-based 3D hand reconstrcution, temporal context}


\maketitle

\section{Introduction}
Interacting hand refers to the left and right hands collaborating together to perform some actions.
The analysis of interacting hands plays an important role in human behavior understanding, thus facilitating the development of many applications, \emph{i.e.,} AR/VR technology, sign language and human-machine interaction, \emph{etc.}.
Recent years have witnessed great success of single hand pose and shape recovery thanks to the development of deep neural networks~\cite{boukhayma20193d, zhou2020monocular, wang2018mask, xu2020improve, sanchez2017robust, liang2014resolving, yang2020seqhand}.
Compared with the single hand case, the reconstruction of pose and shape for both hands is much more challenging due to interfering factors, such as self-occlusion, mutual occlusion and similar textures. 
As a result, most existing single-hand reconstruction methods~\cite{boukhayma20193d, yang2020seqhand, wang2018mask, liang2014resolving} cannot be directly applied to the interacting hand scenario. 

In order to promote the study of interacting hand reconstruction, several  datasets have been made available to the research community~\cite{moon2020interhand2, hasson2019learning}. 
Currently, the research on interacting hand reconstruction has received increasing attention~\cite{mueller2019real,han2020megatrack, moon2020interhand2,lin2021two, fan2021learning, zhang2021interacting, kim2021end}, which can be categorized into two groups based on input modality, \emph{i.e.,} RGB data and depth data. 
In this work, we focus on interacting hand reconstruction from monocular RGB data.
RGB-based methods~\cite{moon2020interhand2, lin2021two, kim2021end} estimate the sparse 3D hand pose for both hands by extracting both hand features from a single color image.
For instance, Zhang \emph{et al.}~\cite{zhang2021interacting} attempt to embed the two-hand context information to reconstruct interacting hand pose and shape from a single image. 
Meng~\emph{et al.}~\cite{meng20223d} propose to segment the masks of the left and the
right hand in a single image and utilize 3D single-hand pose estimator to process both hands separately.
However, these methods rely solely on the limited information available from a single color image in order to extract the single hand pose feature, leading to inaccuracies in estimating 3D joints, particularly in scenarios involving heavy self-occlusion or mutual-occlusion of both hands. 

While the current methods have achieved remarkable success in extracting isolated hand information from a single image, they often overlook the abundant spatial and temporal information present in adjacent frames. 
Interacting hands, in particular, exhibit a strong spatial correlation, which not only minimizes the likelihood of physical interpenetration but also facilitates the precise reconstruction of both hands. 
Additionally, temporal information provides context for the relationship between the hands, which can help overcome the limitations of occlusions in a single frame.

Motivated by the analysis above, we propose a novel framework  that leverages spatial-temporal information to achieve superior reconstruction of interacting hand pose and shape. As shown in Fig.~\ref{fig1}, our framework mainly consists of five modules, \emph{i.e.,} a frame encoder, a multi-scale feature extractor, a temporal encoder, a MANO decoder and an interpenetration detector. 
Given an interacting-hand sequence, the frame image encoder first transforms it into a series of image features. 
Then, using a multi-scale feature extractor, we extract coarse features for both hand and hierarchical image features for each frame. 
Subsequently, the feature sequences of both hands are fed into the temporal encoder and progressively aggregated for representation refinement. 
Finally, the MANO decoder maps the latent inter-hand semantic representation to the 3D hand mesh and pose. 
The interpenetration detector is utilized to detect the collision information for both predicted hands. 
It is worth mentioning that, to alleviate the irrational hand pose and generate realistic interacting hand action, we design a temporal constraint and a physical constraint to guide our framework in learning reasonable hand representation.

The key contributions of this work are summarized as follows:
\begin{itemize}
	\item We propose an interaction-aware framework to explicitly exploit spatial-temporal information for hand reconstruction under the challenging interacting hand scenario.  
	\item We elaborately design a novel temporal framework with the temporal constraint to generate smooth interacting hand motion. Besides, we propose an interpenetration detection module to reconstruct interacting hands satisfying the physical constraint.
	\item Extensive experiments are performed to validate the effectiveness of our proposed framework, which achieves new state-of-the-art performance on two public benchmarks. 
\end{itemize}

\section{Related Work}
In this section, we briefly review several related topics, including 3D interacting hand pose and shape estimation and other hand pose estimation.

\subsection{3D Interacting Hand Estimation} 
Most existing methods conduct inter-hand reconstruction with multi-view cameras \cite{ballan2012motion, han2020megatrack}, a single depth image \cite{mueller2019real, hasson2020leveraging} or tracking strategy \cite{wang2020rgb2hands, smith2020constraining}. 
Based on the input modality, these methods are mainly categorized into depth-based and RGB-based methods. The RGB data and depth data are generally different. RGB images primarily record pixel values for the RGB channels, while depth images primarily contain distance values. 
With the popularity of portable devices, the large-scale RGB data is more readily available than the collection of depth data.

Due to the extensive use of color images, the methods using monocular RGB images~\cite{moon2020interhand2, zhang2021interacting, lin2021two, kim2021end,Li2022intaghand,meng20223d} gradually become the mainstream.
Moon \emph{et al.} \cite{moon2020interhand2} propose the current largest interacting-hand dataset named InterHand2.6M and present an InterNet model. 
InterNet extracts the hand feature for each hand from the single RGB image, then predicts a 2.5D heatmap with a single hand feature and lifts it to 3D sparse joints.
Kim \emph{et al.}~\cite{kim2021end} combine an inter-hand detection network and an inter-hand pose estimation network to estimate the intermediate 2D interacting hand pose, and then lift sparse 3D joints from 2D hand pose positions.
Zhang \emph{et al.}~\cite{zhang2021interacting} leverage two-hand context and 2.5D heatmap to reconstruct interacting hand pose and shape from a single color image. The method consists of two stages, \textit{i.e.,} predicting 2.5D heatmap of each joint as an initial estimation of the interacting hands, then refining the initial estimation with high-resolution features to capture more details.
Meng \emph{et al.}~\cite{meng20223d} propose to convert the issues of interacting hands into the single-hand domain and take advantage of the research progress on the single-hand pose estimation system to predict the joints of both hands respectively. Thus, this method focuses on designing a hand segmentation module to modal the masks of both hands and apply de-occlusion technology to improve pose estimation accuracy.

However, for interacting hand pose and shape reconstruction from RGB images, the aforementioned methods do not explicitly consider the temporal information and spatial information between both hands.
Different from solely relying on a single RGB image, our proposed method explicitly exploits the temporal-spatial information among consecutive frames for reconstructing interacting hands.

\subsection{Other Hand-Related Pose Estimation}
Other hand-related pose estimation mainly involves single hand estimation and hand-object pose estimation. Early 3D single hand pose estimation methods are mainly based on a generative model~\cite{qian2014realtime, sharp2015accurate, tagliasacchi2015robust, tang2015opening}. 
Recently, many works utilize the powerful deep neural networks to estimate the single hand pose~\cite{tompson2014real, guo2017region, zimmermann2017learning, moon2018v2v,liu2021han,cai2018weakly, li2020exploiting,iqbal2018hand, wan2019self, boukhayma20193d, yang2020seqhand, huang2020awr, huang2017ego, xiao22real, jiaoglpose}.
RT-CPR \cite{tompson2014real} first utilizes the deep neural network to localize hand keypoints by estimating 2D heat maps for each joint.
This method is further extended by estimating multi-view 2D \cite{guo2017region}. 
V2V-PoseNet \cite{moon2018v2v} proposes a 3D CNN model which takes voxel input and outputs a 3D heat map for each keypoint. 
In~\cite{wan2019self}, Self-HPE is presented as a self-supervised method that can be trained using only a single depth map. 
In~\cite{huang2020awr}, AWR is proposed as an adaptive weighting regression method to estimate hand joint coordinates in dense representation.
Boukhayma \emph{et al.}~\cite{boukhayma20193d} enhance the model generalization by introducing extra data resources and introducing a hand prior model. 
SeqHand~\cite{yang2020seqhand} incorporates the temporal movement information of isolated hand sequences with the static frame information for better 3D hand pose estimation. 
Hu~\textit{et al.,}~\cite{hu2021hand,hu2023signbert+,zhao2023best} utilize pre-training techniques to predict 3D hand meshes from 2D joints with a hand statistical model in the sign language domain.
Group-HPE~\cite{li2020exploiting} separates the hand joints into different groups by calculating the relationship among them. 
$\mathrm{S}^2\mathrm{HAND}$~\cite{chen2021model} trains a model-based 3D hand reconstruction network with a self-supervised paradigm to alleviate reliance on labeled training data. 
With the success of single hand pose estimation, some works begin to study hand-object pose estimation~\cite{ballan2012motion, wang2013video, tzionas20153d, romero2010hands, hasson2019learning, narasimhaswamy2020detecting, shan2021cohesiv, diffenderfer2021winning, hasson2019learning, fan2021understanding}. Due to the impenetrable nature of hands and objects, many hand-object reconstruction works~\cite{hasson2019learning,yang2021cpf,grady2021contactopt} investigate how to solve the problem of interpenetration between hands and objects. Hasson \textit{et al.}~\cite{hasson2019learning} adopt a contact loss to restrain the distance between hand and object meshes. ContactOpt~\cite{grady2021contactopt} proposes a differentiable contact model to optimize hand pose to achieve desirable contact.

Since the aforementioned hand related reconstruction methods either only focus on the single-hand estimation or rely on the characterization of the hand-object space position, they can not be directly extended to the interacting hand scenarios.
We are the first to consider the spatial and temporal correlation between both hands with interpenetration constraints and temporal smoothness and prove the effectiveness of our proposed method.

\begin{figure*}[t]
	\centering
	\includegraphics[width=1.0\textwidth]{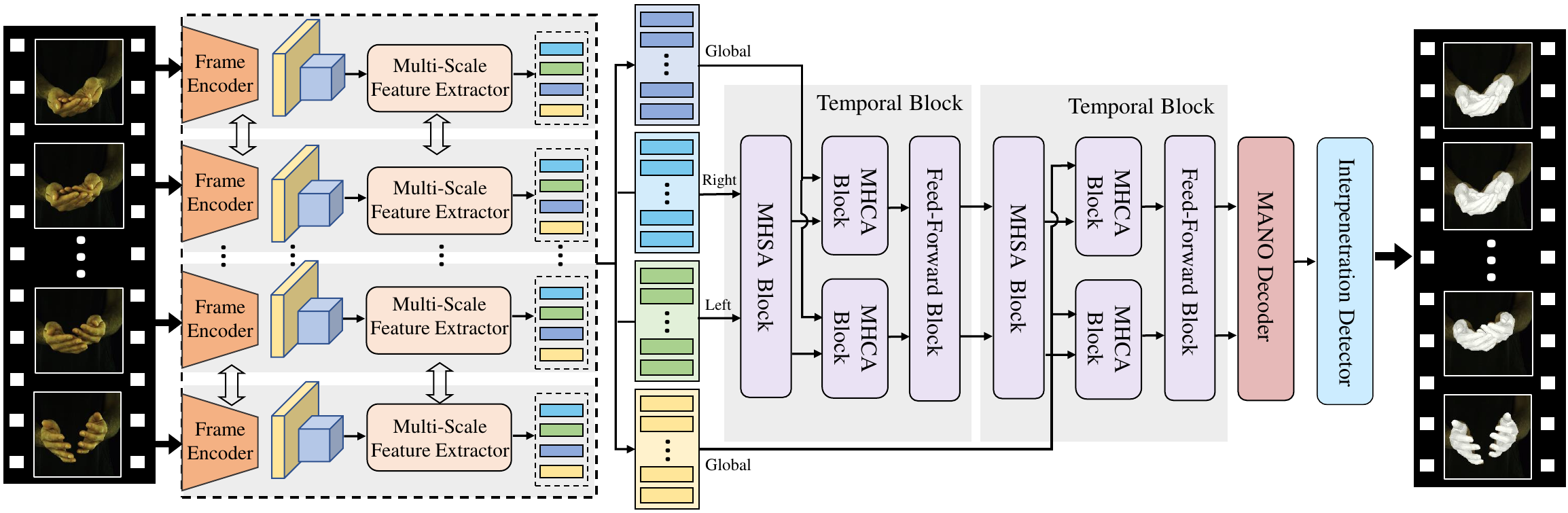} 
	\caption{An illustration of our proposed framework. 
		Given a cropped RGB sequence, the frame encoder first distills image features for each frame. Then the multi-scale feature extractor extracts multi-scale feature sequences for temporal modeling. The temporal encoder refines both hand features with temporal contexts. Finally, the MANO decoder regresses the 3D coordinates of two hand surface vertices, and the interpenetration detector detects whether two hands occur collision during training. The weights of the frame encoder and multi-scale feature extractor are shared among RGB frames.} 
	\label{fig1}
\end{figure*}

\section{Methodology}
As shown in Fig.~\ref{fig1}, our framework consists of a frame encoder, a multi-scale feature extractor, a temporal encoder, a MANO decoder and an interpenetration detector. The temporal encoder is made up of two temporal blocks. 
In the following, we will elaborate each component of our framework. 
After that, we discuss the objective function for training.

\subsection{Framework Architecture}

\textbf{Frame encoder.}  Given an input RGB interacting hand sequence $\mathit{V}=\{\mathit{I}_t\}_{t=1}^T$ of length $\mathit{T}$ from a single person, we extract the image features of the last two layers, \emph{i.e.,} $F_{t, l}^{img} \in \mathbb{R}^{C_{l} \times H_{l} \times W_{l}}, l \in \{1, 2\}$ for each frame $\mathit{I}_t$ using a pretrained ResNet~\cite{he2016deep}. The network weights of the ResNet are shared among all frames. The output of frame encoder is formulated as $\mathbf{F}^{img} = \{\{F_{t, l}^{img}\}_{l=1}^2\}_{t=1}^{\mathit{T}} = \{\mathbb{E}(\mathit{I}_t)\}_{t=1}^{\mathit{T}} ,$
where $\mathbb{E}(\cdot)$ denotes the frame encoder, and $\mathbf{F}^{img}$ represents the set of image features extracted from the frame encoder.

\begin{figure}[t]
	\centering
	\includegraphics[width=0.7\textwidth]{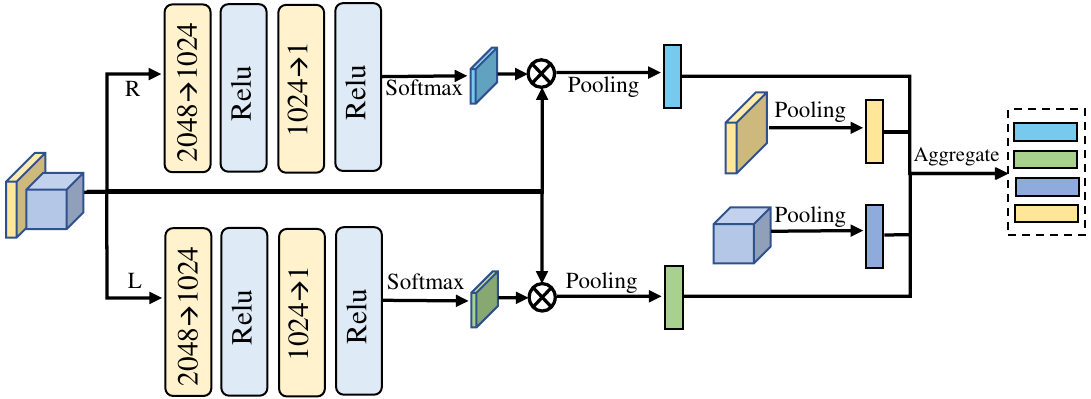}
	\caption{Illustration of the multi-scale feature extractor in our proposed framework.}
	\label{extractor}
\end{figure}

\textbf{Multi-scale feature extractor.} The set of image features $\mathbf{F}^{img}$ includes the static semantics of both hands. In order to extract useful information related to interacting hands, we propose the multi-scale feature extractor, as shown in Fig.~\ref{extractor}. We first utilize a hand-chirality extractor to extract a single hand feature for each frame. It predicts a weight map $w_t^h \in \mathbb{R}^{H_1 \times W_1}$ for each hand that extracts single hand feature $f_t^h \in \mathbb{R}^{C_1}$ from the image feature $F_{t, 1}^{img}$, which is formulated as follows, 
\begin{equation}
	\label{equ:Hand_Extractor_1}
		w_t^h = \mathbb{HE}^h(F_{t, 1}^{img}), \quad f_t^h = \mathbb{P}_{avg}(w_t^h \otimes F_{t, 1}^{img}), \quad h \in \{left, right\} ,
\end{equation}
where $h$ represents left or right hand side, and $\mathbb{HE}^h(\cdot)$ denotes hand chirality extractor for each hand side.  $\otimes$ and $\mathbb{P}_{avg}$ indicates element-wise multiplication and average pooling operations, respectively. We also consider that due to the similar texture and occlusion between both hands, directing reconstruction interacting hand mesh from a single hand feature maybe have difficulty in guaranteeing plausible predicted results. To this end, we extract the hierarchical global features, \emph{i.e.,} $f_{t, 1}^{global} \in \mathbb{R}^{C_1}$ and $ f_{t, 2}^{global} \in \mathbb{R}^{C_2}$ by utilizing average pooling operation on image features $F_{t, 1}^{img}$ and $F_{t, 2}^{img}$ for each frame, respectively. Finally, we collect four multi-scale features for each frame, \emph{i.e.,} $f_t^{right}, f_t^{left}, f_{t,1}^{global}$ and $f_{t,2}^{global}$. To facilitate the introduction below, we split them into four feature sequences, denoted as $F^{right}, F^{left}, F_{1}^{global} \in \mathbb{R}^{T \times C_1}$ and $F_2^{global} \in \mathbb{R}^{T \times C_2}$, respectively.

\textbf{Temporal encoder.} 
Given the above feature sequences, we design the temporal encoder to exploit rich context cues and progressively merge the temporal information. The temporal encoder is composed of temporal blocks motivated by~\cite{vaswani2017attention}. Each temporal block is formed by three submodules: a multi-head self-attention~(MHSA) block, two multi-head cross-attention~(MHCA) blocks and a feed-forward block. 
Two hand feature sequences are first cascaded and fed into MHSA block. MHSA block is utilized to capture the temporal context to compensate insufficient information in a single frame caused by interfering factors. Meanwhile, the MHSA block could also implicitly model hands interacting context for both hand sequences, which is formulated as follows,
\begin{equation}
	\label{equ:Temporal encoder}
		F_{in, 1}^{k} = \mathit{Concat}(F^{right}, F^{left}), \quad
		F_{out, 1}^{k} = \mathrm{MHSA}(F_{in,1}^{k}), 
\end{equation}
where $F_{in, 1}^{k} \in \mathbb{R}^{2T \times C_k}$ and $F_{out, 1}^{k}  \in \mathbb{R}^{2T \times C_k}$,  denotes the input and output feature sequence of the MHSA block~(the $1^{st}$ submodule) in the $k$-th temporal block, respectively. $F_{out, 1}^{k}$ can be split into two parts, \emph{i.e.,} $F_{out, 1}^{left, k}$ and $F_{out, 1}^{right, k}$.
Through the MHSA block, each hand feature refines itself with hand temporal information. However, single hand feature only contains local information extracted from the frame image feature. In order to convey global context into local hand feature sequence, we utilize multi-head cross-attention (MHCA) block to merge hierarchical global feature sequence into both hand feature sequences respectively, which is formulated as follows,

\begin{equation}
    \label{equ:Cross Attention}
    F_{out, 2}^{h,k} = \mathrm{MHCA}( F_{out, 1}^{h,k}, F_{k}^{global})
\end{equation}
where $h$ represents the hand side, \emph{i.e.,} right and left. Finally, both hand feature sequences are fed into feed-forward block with a dimensionality reduction scheme for the next temporal block. Note that, we totally utilize two temporal blocks in our proposed framework.

\textbf{MANO decoder.} 
We utilize the fully differentiable MANO hand model~\cite{romero2017embodied} as the decoder to represent the pose and shape of interacting hands in the camera coordinate system. As a statistical model, the MANO model introduces hand prior to constrain the distribution of possible inter-hand poses. The irrational predicted hand poses are implicitly filtered out during its mapping. For the outputs of the temporal encoder, $F_{out}^{right}, F_{out}^{left} \in \mathbb{R}^{T \times C_3}$, we first utilize an MLP as latent semantic extractor to regress the MANO shape parameters $\hat{\bm{\beta}}^{right}, \hat{\bm{\beta}}^{left} \in \mathbb{R}^{T \times 10}$, pose parameters $\hat{\bm{\theta}}^{right}, \hat{\bm{\theta}}^{left} \in \mathbb{R}^{T \times 10}$ for both hand sequences, respectively. Then the sequence of hand surface mesh vertices $\mathbf{M}^{h} \in \mathbb{R}^{\mathit{T} \times 778 \times 3}$ for each hand side can be deformed with $\hat{\bm{\theta}}^{h}$ and $\hat{\bm{\beta}}^{h}$.
Given the ground-truth relative translation $\mathbf{c} \in \mathbb{R}^{T \times 1 \times 3}$ between the right and left hand, we could transform left hand mesh vertices into the right hand coordinate system, which is denoted as$\mathbf{M}^{left}_{trans}$ and $\mathbf{M}^{right}_{trans}$.
For each single hand, the predicted hand mesh vertices $\widetilde{V}_{trans, t}^h \in \mathbb{R}^{778 \times 3}$ in $\mathbf{M}^{h}_{trans} $, $t \in \{1, \cdots, \mathit{T}\}$ and $h \in \{left, right\}$, provide a better display of hand pose than sparse skeleton joints. Furthermore, to keep consistent with supervised labels, we derive 21 3D joints $\widetilde{J}_{t}^h \in \mathbb{R}^{21 \times 3}$ from $\widetilde{V}_{trans, t}^h$.

\textbf{Interpenetration detector.} 
So far, the prediction of both hands does not leverage the constraints that guide hands interacting in the physical world. Concretely, it does not explicitly account for the spatial information that both hands can not interpenetrate each other in reality. To better constrain the correct posture of both hands under complex interaction scenarios, we propose an inter-hand penetration detector to detect the predicted interacting hand vertices that are inside each other. We define the distances between points in the following. $d(v, V) = \mathrm{Inf}_{w \in V} || v - w ||_{2}$ denotes the distances from point to set, and $d(C, V) = \mathrm{Inf}_{v \in C} d(v, V)$ denotes the distances from set to set.
Given a paired interacting hand vertices, $V^{right}$ and  $V^{left}$, we first calculate the distances from all vertices of one hand to the other hand, denoted as $d(V^{right}, V^{left})$ and $d(V^{left}, V^{right})$, respectively. Then we cast a ray from each hand vertice and count the number of times it intersect the other hand mesh to determine whether it is inside or outside the predicted hand mesh, which is formulated as follows,
\begin{equation}
	\label{equ:penetration}
		\mathrm{Inter}(V^{h}) = \mathit{CastRay}(V^{h}, Mesh(V^{\hat{h}}, F^{\hat{h}})), \quad
		h, \hat{h} \in \{right, left\}
\end{equation}
where $\mathrm{Inter}(\cdot)$ denotes whether interpenetration occurs at each point in hand meshes. The value of $\mathrm{Inter}(\cdot)$ is only taken from the set of $\{1, 0\}$, which indicates that interpenetration happens or not, respectively. $Mesh(\cdot)$ utilizes hand vertices coordinates and the connection relationship of hand vertices to construct a waterproof hand mesh. $h$ and $\hat{h}$ denotes both hand sides, respectively. According to the interpenetration detection results, we explicitly detect the penetration regions and design physical interacting constraint to reduce them.

\subsection{Objective Function}
During training, we introduce objective functions to supervise our framework outputs. Specifically, we design loss functions for temporal and spatial constraints. 

\textbf{Temporal smooth loss.} During the hand motion, each hand pose sequence should be continuous and smooth. 
In order to enforce the temporal smoothness of interacting hand movement, we adopt temporal smooth loss during the training procedure, which is formulated as follows,
\begin{equation}
	\label{equ:temporal loss}
	\begin{split}
		\mathcal{L}_{smooth} = \sum\limits_{h \in \{l, r\}} \sum\limits_{t=2}^{\mathit{T}} \lVert \widetilde{J}_{t}^h - \widetilde{J}_{t-1}^h \rVert_2,  \\
	\end{split}
\end{equation}
where $\widetilde{J}_{t}^h$ denotes the predicted hand joint pose in the $t$-th frame, and $h$ denotes the hand side.

\textbf{Interpenetration loss.} We define a interact loss to penalize both hands interpenetration. Moreover, we utilize a common penalization function $l_{\alpha}(x) = \alpha \cdot \mathrm{tanh}(\frac{x}{\alpha})$, where $\alpha $ is the penetration distance, which we empirically set to $2cm$ in all experiments. Thus, the interact loss is formulated as follows,
\begin{equation}
	\label{equ:inter loss}
	\begin{split}
		\mathcal{L}_{inter} = \sum\limits_{h \in \{l, r\}} \sum\limits_{t=1}^{\mathit{T}} (\mathrm{Inter}(\widetilde{V}_{trans, t}^{h}) \cdot l_{\alpha}(d(\widetilde{V}_{trans, t}^{h}, \widetilde{V}_{trans, t}^{\hat{h}}))),  \\
	\end{split}
\end{equation}
where $\widetilde{V}_{trans, t}^{h}$ denotes the predicted hand mesh vertices for each hand side.

\textbf{Joint loss.} 
To better constrain the pose of the generated interacting hand under a real situation, we add 3D joint loss $\mathcal{L}_{3D}$ during the training procedure, which is formulated as follows,
\begin{equation}
	\label{equ:joint loss}
	\begin{split}
		\mathcal{L}_{joint} &= \frac{1}{\mathit{T}}\sum\limits_{h \in \{l, r\}} \sum\limits_{t=1}^{\mathit{T}} \mathbf{1}(t) \cdot\lVert J_{t}^h - \widetilde{J}_{t}^h \rVert_2,  \\
	\end{split}
\end{equation}
where $J$ and $\widetilde{J}$ denote the groundtruth and the predicted joints from our framework, and $\mathbf{1}(t)$ is an indicator function that is 1 if the ground truth 3D joints in the $t-$th frame are labeled and 0 otherwise. 

\textbf{MANO loss.} 
We utilize MANO loss to enforce the hand shape parameters. Meanwhile, due to the symmetry of both hands of a subject, we also keep the hand shape consistency of both hands with L2 distance. The mano loss is formulated as follows,
\begin{equation}
	\label{equ:mano loss}
		\mathcal{L}_{mano} = \sum\limits_{h \in \{l, r\}}\sum\limits_{t=1}^{\mathit{T}} \mathbf{1}(t)\cdot \lVert \bm{\beta}_t^{h} -  \hat{\bm{\beta}}_t^{h} \rVert_2  + \lambda_{consist} \sum\limits_{t=1}^{\mathit{T}}  \lVert \bm{\beta}_t^{right} -  \hat{\bm{\beta}}_t^{left} \rVert_2^{2},
\end{equation}
where $\lambda_{consist}$ denotes weighting factor of hand shape consistency, which is set to 1.0 in our settings.

\textbf{Regularization Loss.} To generate plausible inter-hand poses and accelerate network convergence,  the regularization loss is added by constraining the magnitude of the MANO model's inputs, which is formulated as follows,
\begin{equation}
	\label{equ:regular loss}
	\mathcal{L}_{reg} = \sum\limits_{h \in \{l, r\}} \lVert \bm{\hat{\theta}}^{h} \rVert_2^2+ \lambda_{\beta} \lVert \bm{\hat{\beta}}^{h} \rVert_2^2 ,
\end{equation}
\noindent where $\lambda_{\beta}$ denotes weighting factor of hand shape parameters, which is set to 0.1 in our settings.

The total loss function of our framework is defined as follows,
\begin{equation}
	\label{equ:total loss}
	\mathcal{L}_{total} = \mathcal{L}_{smooth} + \lambda_{j} \mathcal{L}_{joint} +  \lambda_{i} \mathcal{L}_{inter} + \lambda_{m} \mathcal{L}_{mano} + \lambda_{r} \mathcal{L}_{reg},
\end{equation}
where $\lambda_{j}, \lambda_{i}, \lambda_{m}$, and $\lambda_{r}$ are the loss weights, and they are set to 100, 10.0, 1.0 and 0.1, respectively. 

\section{Experiment}
In this section, we first describe the experiment setup, including implementation details, datasets and evaluation metrics. 
Next, we compare our method with previous approaches and conduct an ablation study to validate the effectiveness of our method. 
Finally, we make qualitative analysis to demonstrate the superiority of our method.

\subsection{Experiment Setup}
\label{exp setup}	\textbf{Implementation details.} We use the ResNet-50~\cite{he2016deep} as the frame encoder, which is pre-trained on the ImageNet~\cite{deng2009imagenet} and jointly trained with the whole framework. The size of input frame is cropped to $256 \times 256$ pixels, and the output dimensionalities of ResNet-50 last two layers $C_1, H_1, W_1$, and $C_2, H_2, W_2$ are set to 2048, 8, 8, and 1024, 16, 16, respectively. The sequence length $\mathit{T}$ is set to 10. In our experiment, all the models are implemented in PyTorch \cite{paszke2019pytorch} platform and trained on NVIDIA RTX-3090. We use Adam optimizer \cite{kingma2014adam} with a minibatch size of 32 per GPU. The weight decay and momentum are set to 1e-4 and 0.9, respectively. The learning rate is initialized to 0.001 and reduced by a factor of 0.1 every 10 epochs. We train our framework for 30 epochs in total.

\textbf{Datasets.} We train our proposed network and evaluate the performance on two public inter-hand reconstruction datasets, InterHand2.6M \cite{moon2020interhand2} and HIC \cite{hasson2019learning}. Since our method aims to reconstruct the mesh of interacting hands from video-based RGB images, we need the available inter-hand datasets which provide inter-hand mesh annotations. As far as we know, InterHand2.6M and HIC are the only existing real-world datasets that satisfy the above requirement.

InterHand2.6M \cite{moon2020interhand2} provides data of two different frame rates, including 5 FPS and 30 FPS. Since our method focuses on temporal sequence, we use the version of 30 FPS, consisting of 2,202,883 for training and 1,717,779 for testing. The resolution of all frames is fixed to $334 \times 512$. In terms of the size of this dataset, InterHand2.6M is the most challenging and largest interacting hand dataset.

HIC \cite{hasson2019learning} consists of single-hand frames, interacting hand frames and hand-object interaction frames. We select the interacting hand part, containing a total of 1307 frames with 7 video sequences. In this subset, the resolution of all frames is fixed at $640 \times 480$.  
Due to the less number of video sequences, we achieve the extension of the subset by sampling a fixed number $L$ of frames from each original video. Finally, we set $L$ to 20 and get 806 sequences for training, and 375 sequences for testing.

\textbf{Evaluation metrics.} In our experiment, we report the single-frame and temporal evaluation metrics. For the single-frame evaluation, we use mean per joint position error (MPJPE), Procrustes-aligned MPJPE (PA-MPJPE) and mean per-vertex position error (MPVPE). All the position errors are measured in millimeter ($mm$) between the predicted and ground-truth 3D joint positions after aligning the root joint following~\cite{zhang2021interacting}. 
We also adopt the Percentage of Correct Keypoints (PCK) and Area Under the Curve (AUC) (0-50) mm of PCK curve over different error thresholds to evaluate the accuracy of 3D hand pose estimation. 
For temporal evaluation, we report acceleration error~(Accel\_E) proposed in HMMR \cite{kanazawa2019learning}. The acceleration error computes the difference between the estimated and ground-truth acceleration of each joint in ($mm/s^2$). 
For penetration evaluation, we report the mean maximum penetration depth~(MMPD) measured in millimeter~($mm$). If the two hands collide, the penetration depth is the maximum distance from the right hand mesh vertices to the left hand's surface. In the absence of collision, the penetration depth is 0. In the qualitative results, we align the root joint
of each hand with the ground truth for visualization.

\begin{table*}[t]
	\normalsize
	\tabcolsep=1.5pt
	\renewcommand\arraystretch{1.0}
	\begin{center}
		\resizebox{1.0\linewidth}{!}{
			\begin{tabular}{l|ccccc|ccccc}
				\hline
				\multirow{2}{*}{Methods}      &  \multicolumn{5}{c|}{\textbf{InterHand2.6M}} & \multicolumn{5}{c}{\textbf{HIC}} \\ 
				\cline{2-11}
				& Accel\_E & MMPD & MPJPE  & MPVPE  & PA-MPJPE & Accel\_E & MMPD & MPJPE  & MPVPE  & PA-MPJPE \\ \hline \hline
				InterNet~\cite{moon2020interhand2} & 13.46 & -- & 17.81 & -- & 9.21 & 7.40 & -- & 9.83 & -- & 6.74 \\
				HPS~\cite{boukhayma20193d} & 8.14 & 7.96 & 19.40 & 19.78 & 9.51 & 6.26 & 1.95 & 8.11 & 5.57 & 7.09  \\
				ITH-3D~\cite{zhang2021interacting} & 6.01 & 7.67 & 14.45 & 14.74 & 8.64 & 3.36 & 2.97 & 7.29 & 8.76 & 5.25\\
                    DIGIT~\cite{fan2021learning}  & 6.78 & -- & 14.57 & -- & 8.53 & -- & -- & -- & -- & -- \\
				SeqHand~\cite{yang2020seqhand} & 6.69 & 8.11 & 18.20 & 19.46 & 9.24 & 2.02 & 2.85 & 7.64 & 8.15 & 4.38 \\
				Ours & \textbf{3.70} &  \textbf{4.22} & \textbf{13.25} &  \textbf{13.91} &  \textbf{8.12} & \textbf{1.87} & \textbf{1.73} & \textbf{6.08} &  \textbf{6.90} &  \textbf{3.73}  \\ \hline
			\end{tabular}
		}
		\caption{Comparison with state-of-the-art methods on InterHand2.6M~\cite{moon2020interhand2} and HIC~\cite{hasson2019learning} datasets. Since InterNet \cite{moon2020interhand2} and DIGIT~\cite{fan2021learning} only predicts sparse 3D joint coordinates, we do not report the MPVPE and MMPD for this method. HPS \cite{boukhayma20193d} and SeqHand \cite{yang2020seqhand} are shape reconstruction methods. For all the five evaluation metrics, a small number indicates a better performance.The best performance under each metric is highlighted in bold.}
		\label{table:SOTA}
	\end{center}
		\vspace{-2.0em}
\end{table*}

\begin{figure*}[t]
	\centering
	\includegraphics[width=\textwidth]{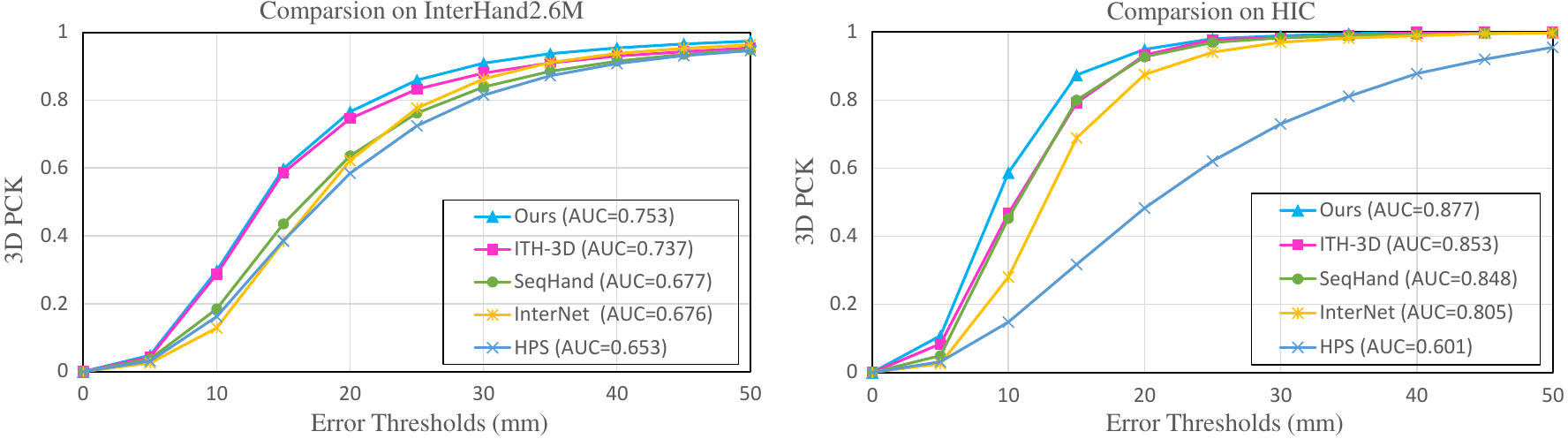}
	\caption{Quantitative comparison of our proposed method with the previous methods on InterHand2.6M~\cite{moon2020interhand2} and HIC~\cite{hasson2019learning} datasets.  The horizontal axis indicates the error threshold for interacting hand, while the vertical axis indicates the 3D Percentage of Correct Keypoints (3D PCK). 
	}
	\label{fig2}
\end{figure*}

\begin{figure*}[!h]
	\centering
	\includegraphics[width=\textwidth]{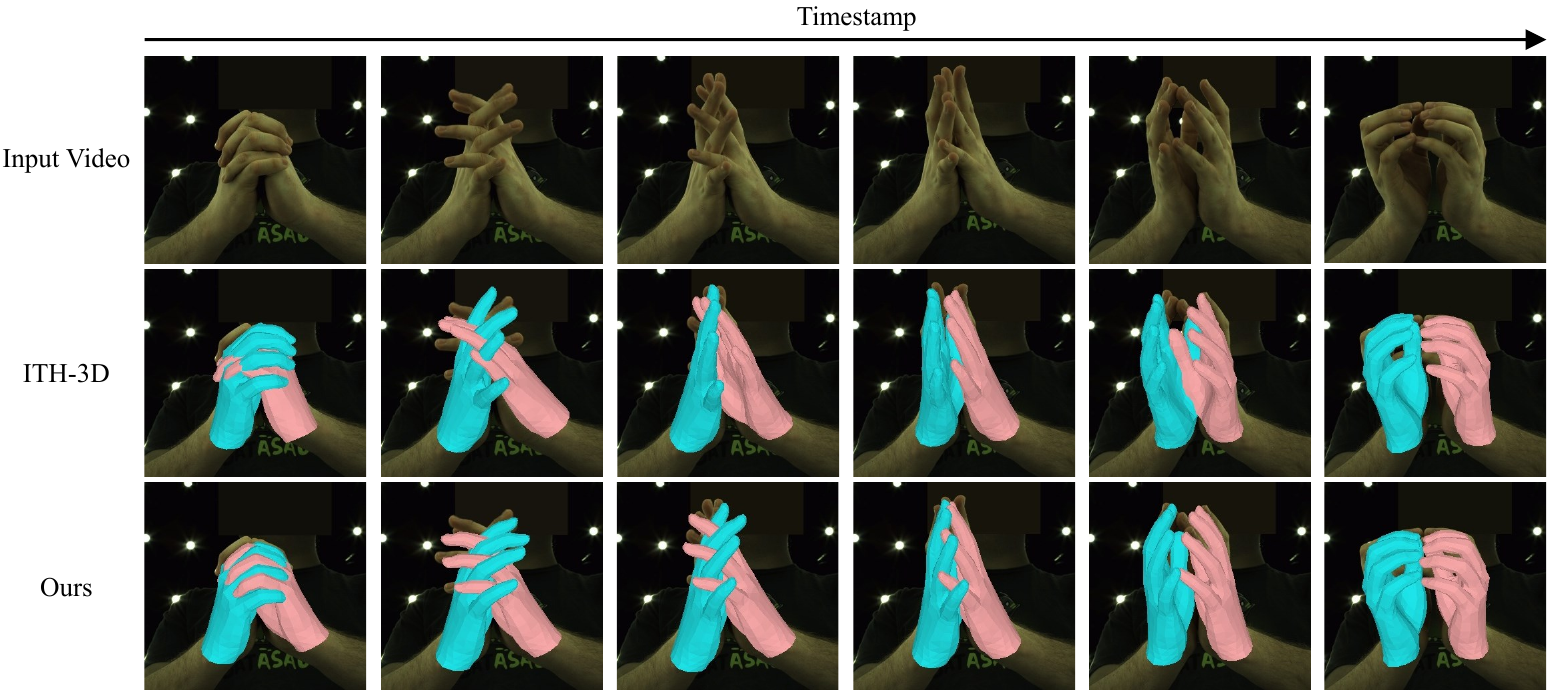}
	\caption{Comparison of qualitative results between the state-of-the-art method, ITH-3D~\cite{zhang2021interacting} and our proposed method. Our method produces more accurate and smooth interacting hand poses, while ITH-3D~\cite{zhang2021interacting} produces more collisions between both hands and worse hand poses.}
	\label{fig4}
\end{figure*}

\subsection{Comparison with State-of-the-art Methods}
As shown in Tab.~\ref{table:SOTA}, we compare our approach with previous frame-based and temporal methods. 
For frame-based methods, ITH-3D~\cite{zhang2021interacting} is the state-of-the-art interacting hand pose reconstruction method. InterNet~\cite{moon2020interhand2} is another method that predicts sparse 3D joints for interacting hands. HPS~\cite{boukhayma20193d} is an isolated hand reconstruction method with the hand statistical model MANO~\cite{romero2017embodied}. 
For temporal methods, SeqHand \cite{yang2020seqhand} is proposed for isolated hand pose sequence estimation. 

On InterHand2.6M~\cite{moon2020interhand2} and HIC~\cite{hasson2019learning} datasets, our proposed method significantly outperforms all the single hand approaches in all evaluation metrics, presumably because they do not consider heavy hand occlusions and confuse about the similar appearance of both hands. Compared with the two-hand pose estimation method InterNet~\cite{moon2020interhand2}, our method reduces the MPJPE by 25.6\% (17.81$\rightarrow$13.25),  38.14\% (9.83$\rightarrow$6.08) on InterHand2.6M and HIC, respectively. Our method also gets better performance compared with ITH-3D~\cite{zhang2021interacting}, especially decreasing the acceleration error~(Accel\_E) and mean maximum penetration depth~(MMPD) by 38.4\%~(6.01$\rightarrow$3.70) and 45.0\%~(7.67$\rightarrow$4.22).  ITH-3D only utilizes a single frame to reconstruct both hand pose and shape, which lacks temporal smoothness. Our method leverages static frame information and temporal information, and successfully reduces the acceleration error by keeping the temporal consistency. It is worth noting that our proposed method significantly reduces MMPD attributed to the interpenetration constraint. The above performance validates the effectiveness of our proposed method.

As shown in Fig.~\ref{fig2}, we visualize the Percentage of Correct Keypoints on 3D joints (3D PCK) for interacting hand. For 3D PCK, a large number indicates a better performance. Compared with previous methods, \emph{i.e.,} InterNet \cite{moon2020interhand2}, HPS \cite{boukhayma20193d}, SeqHand \cite{yang2020seqhand} and ITH-3D~\cite{zhang2021interacting}, our method generates more accurate 3D joints on both datasets.    

Fig.~\ref{fig4} further shows the qualitative comparison with the recent model-based two-hand reconstruction SOTA method, ITH-3D~\cite{zhang2021interacting} on InterHand2.6M~\cite{moon2020interhand2}. Given a sequence of interacting hand RGB images, our method recovers significantly better hand pose and shape. Due to lacking the interpenetration constraint of both hands, the collision of both hands seriously influences the reconstruction results. Different from this method, our method produces plausible hand meshes satisfying physical constraints. 
Moreover, despite no explicit contact restraints, the reconstruction of interacting hands still fosters increased contact. We think that the proposed temporal smooth loss plays a vital role in maintaining temporal coherence in both hand poses, constraining them not to move away from each other arbitrarily. The collaboration of our designed objectives could reconstruct interacting hands more smoothly and accurately.
	
\begin{table}[!h]
	\footnotesize
	\tabcolsep=2.5pt
	\renewcommand\arraystretch{1.0}
	\begin{center}
		\resizebox{0.7\textwidth}{!}{
			\begin{tabular}{l|ccccc}
				\hline
				\multirow{2}{*}{}      &  \multicolumn{5}{c}{\textbf{InterHand2.6M}} \\ 
				\cline{2-6}
				& Accel\_E & MMPD & MPJPE  & MPVPE  & PA-MPJPE \\ \hline \hline
				Baseline & 8.14 & 8.87 & 18.49 & 19.07 & 8.96 \\ 
				Baseline + TE & 4.33 & 8.34 & 14.61 & 15.14 & 8.26\\
				Baseline + TE + PC & 4.13 & \textbf{3.82} & 13.41 & 14.11 & 8.43\\
				Baseline + TE + PC + TC & \textbf{3.70} & 4.22 & \textbf{13.25} & \textbf{13.91} & \textbf{8.12} \\
				\hline
			\end{tabular}
		}
		\caption{Ablation study of our network on InterHand2.6M. `TE' denotes the temporal encoder. `TC' and `PC' denote the temporal constraint and interpenetration constraint, respectively. For all the five evaluation metrics, a small number indicates a better performance. The best performance under each metric is highlighted in bold.}
		\label{table:Ablation}
	\end{center}
		\vspace{-1.5em}
\end{table}

\begin{figure*}[t]
	\centering
	\includegraphics[width=0.9\textwidth]{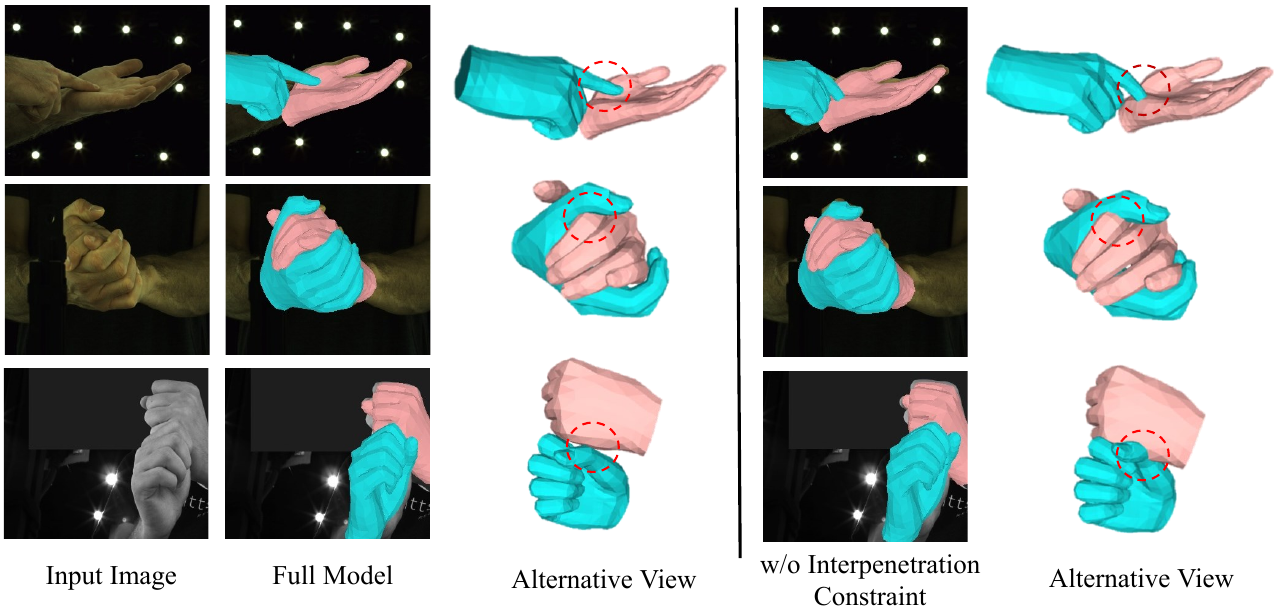}
	\caption{Qualitative ablation study on InterHand2.6M. `w/o Interpenetration Constraint' means removing the spatial information from the full model. The dashed red circles in alternative view display the interpenetration region between both hands. The result shows the effectiveness of spatial information.}
	\label{fig3}
\end{figure*}

\subsection{Ablation Study}
We conduct ablation experiments on InterHand2.6M~\cite{moon2020interhand2} to validate the effectiveness of each component in our framework.

\textbf{Baseline.} We train a baseline model only using the output of the multi-scale feature extractor. As shown in the first row of Tab.~\ref{table:Ablation}, the baseline result shows poor performance and inaccurate interaction reconstruction without temporal context. 

\begin{table}[!h]
	\footnotesize
	\tabcolsep=2.5pt
	\renewcommand\arraystretch{1.0}
	\begin{center}
		\resizebox{0.7\textwidth}{!}{
			\begin{tabular}{l|ccccc}
				\hline
				\multirow{2}{*}{}      &  \multicolumn{5}{c}{\textbf{InterHand2.6M}} \\ 
				\cline{2-6}
				& Accel\_E & MMPD & MPJPE  & MPVPE  & PA-MPJPE \\ \hline \hline
				Ours & \textbf{3.70} &  \textbf{4.22} & \textbf{13.25} & \textbf{13.91} & \textbf{8.12} \\
				Ours~(w/o MANO)  & 5.12 & 6.37 & 14.45 & 14.73 & 8.37 \\
				\hline
			\end{tabular}
		}
		\caption{Ablation study of our network on InterHand2.6M.  ``w/o MANO" denotes that our proposed framework is without the MANO loss and regularization loss. For all the five evaluation metrics, a small number indicates a better performance. The best performance under each metric is highlighted in bold.}
		\label{table:MANO loss}
	\end{center}
		\vspace{-1.5em}
\end{table}

\begin{figure*}[!t]
	\centering
	\includegraphics[width=0.95\textwidth]{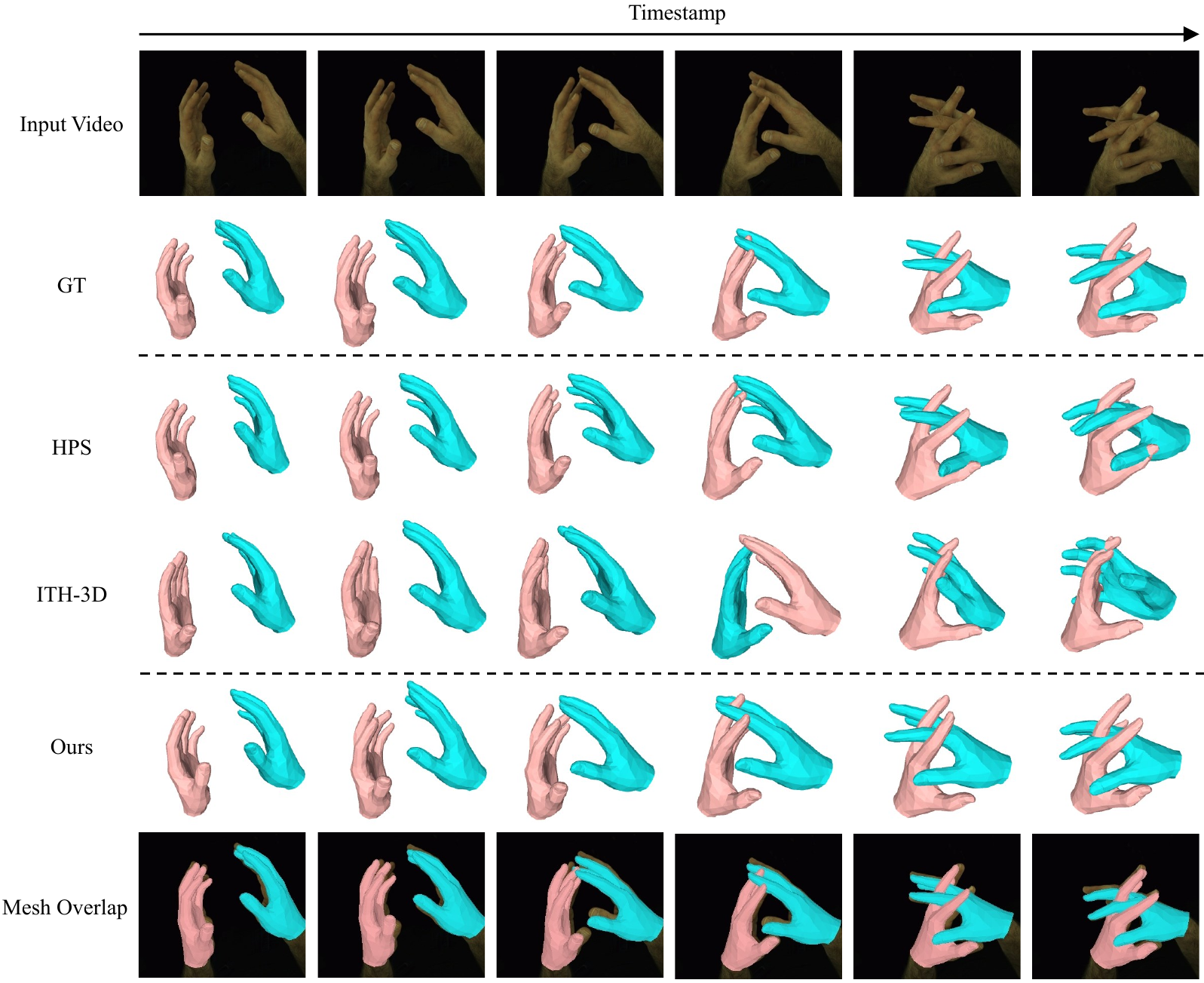}
	\caption{Qualitative comparison of our proposed method with the state-of-the-art methods. Compared with HPS~\cite{boukhayma20193d} and ITH-3D~\cite{zhang2021interacting}, the reconstruction results of our proposed method are closer to the ground truth. ``GT" denotes that the ground truth of interacting hands. }
	\label{fig:comp_with_others}
\end{figure*}

\textbf{Temporal information.} Based on `Baseline', we first add the temporal encoder~(+TE) to capture the context information to alleviate the ambiguities caused by occlusion and similar texture. By modeling the temporal context with the temporal blocks, we achieve significant performance improvement, especially by reducing MPJPE nearly 4mm. Further, we add the temporal smooth constraint~(+TC) to supervise our framework and it achieves better results under all metrics.

\textbf{Spatial information.} Based on `Baseline+TE', we introduce spatial information into our framework with interpenetration constraint~(+PC) between both hands. The result demonstrates the effectiveness of spatial information.
We also show the qualitative result in Fig.~\ref{fig3}. Since both hands usually exists strong interaction, the collision of both hands are inevitable. We display the potential collision region of interacting hands by the alternative view with the dashed red circles. Compared with the results of ``w/o Interpenetration Constraint", our full model generates more plausible results satisfying physical constraints. Therefore, the spatial information is important to the reconstruction of interacting hands.

\textbf{MANO Constraint.} The MANO loss and regularization loss are employed to supervise the predicted parameters of the hand statistical model. Therefore, we conduct an ablation experiment on the constraint of the hand statistical model MANO, and report the results in Tab.~\ref{table:MANO loss}. It is observed that the MANO loss and regularization loss could better constrain the predicted range of hand statistical model parameters and reconstruct more accurate interacting hands.

\subsection{More Comparison and Visualization}
We will show more quantitative and qualitative results, including our proposed method and comparison with other methods, in order to better illustrate the superiority of our method.

\begin{table}[!t]
	\footnotesize
	\tabcolsep=10pt
	\begin{center}
		\resizebox{0.4\linewidth}{!}{
			\begin{tabular}{c|ccccc}
				\toprule
				Methods& Flops  & Params   \\ \midrule
				ITH-3D~\cite{zhang2021interacting} & 32.8G &  141.3M  \\
				Ours & \textbf{16.5}G & \textbf{55.4}M  \\
				\bottomrule
			\end{tabular}
		}
	\end{center}
	\caption{The comparison of the complexity between ITH-3D and our proposed method. ``Flops" represents the abbreviation of floating point operations, and ``Params" represents the abbreviation of the number of model parameters. These two metrics are both utilized to measure the complexity of the model. The lower the value represents the lighter of the model and the smaller computation.}
	\label{flops}
\end{table}

\begin{figure*}[!t]
	\footnotesize
	\centering
	\includegraphics[width=\textwidth]{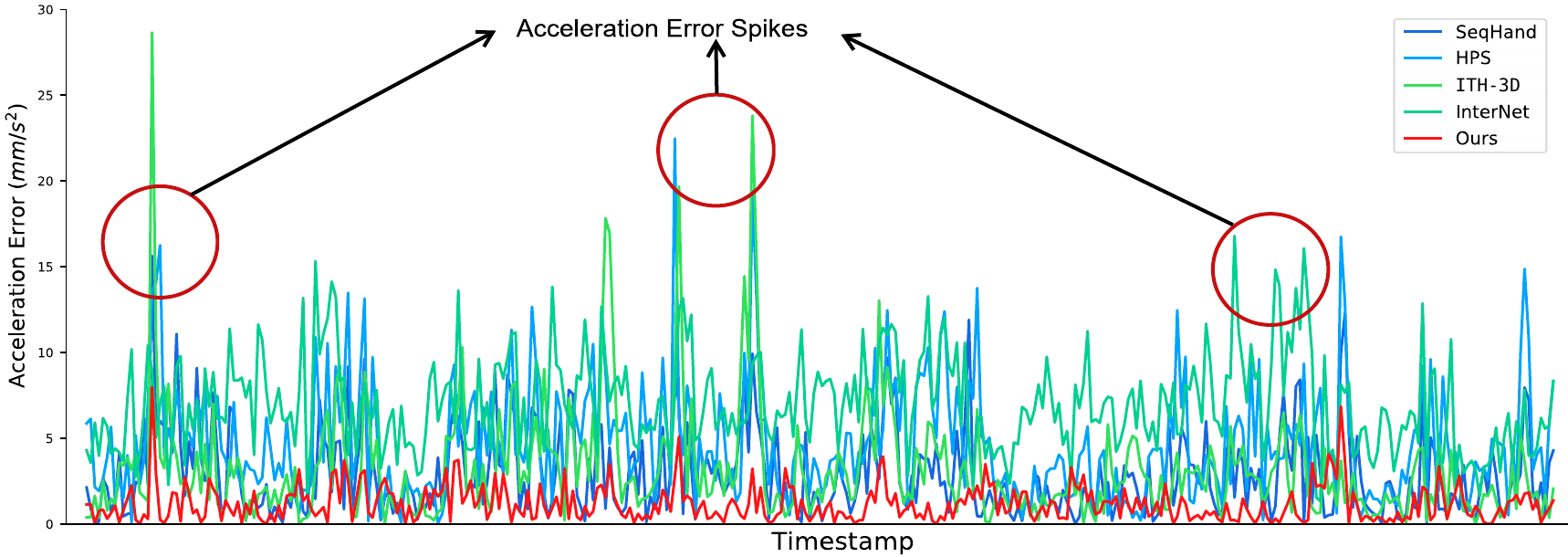} 
	\caption{Comparison of the acceleration errors between our approach and other methods, \emph{i.e.,} InterNet~\cite{moon2020interhand2}, ITH-3D~\cite{zhang2021interacting}, HPS~\cite{boukhayma20193d} and SeqHand~\cite{yang2020seqhand}. The red circles highlight several acceleration error spikes in the test video. Compared with previous methods, our proposed method shows clearly lower acceleration errors along the time step.
	}
	\label{accel}
        \vspace{-1.0em}
\end{figure*}

\textbf{Complexity comparison.} We further compare the complexity of flops and parameters between our proposed method and ITH-3D~\cite{zhang2021interacting} in Tab.~\ref{flops}. It is observed that compared with ITH-3D~\cite{zhang2021interacting}, our approach has only half the computational cost and one-third of the model parameters.

\begin{figure*}[t]
	\centering
	\includegraphics[width=1.0\textwidth]{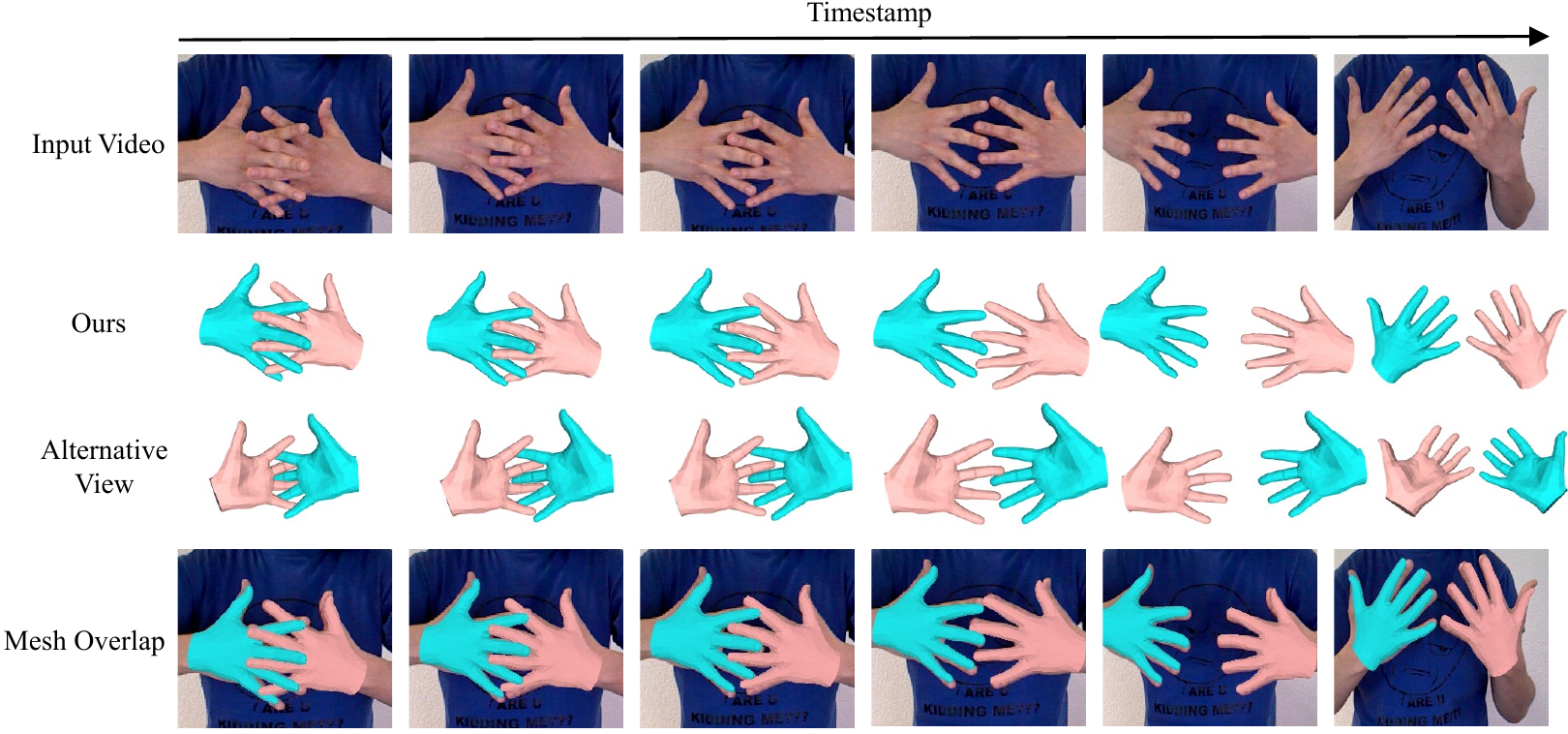}
	\caption{Qualitative results of interacting hand reconstruction of our proposed method on the HIC~\cite{hasson2019learning} dataset.
		For clarification, we visualize the mesh with the two views.
		Our method achieves high-quality reconstruction performance under different levels of occlusion.}
	\label{fig:HIC}
	\vspace{-1.5em}
\end{figure*}
\begin{figure*}[t]
	\centering
	\includegraphics[width=1.0\textwidth]{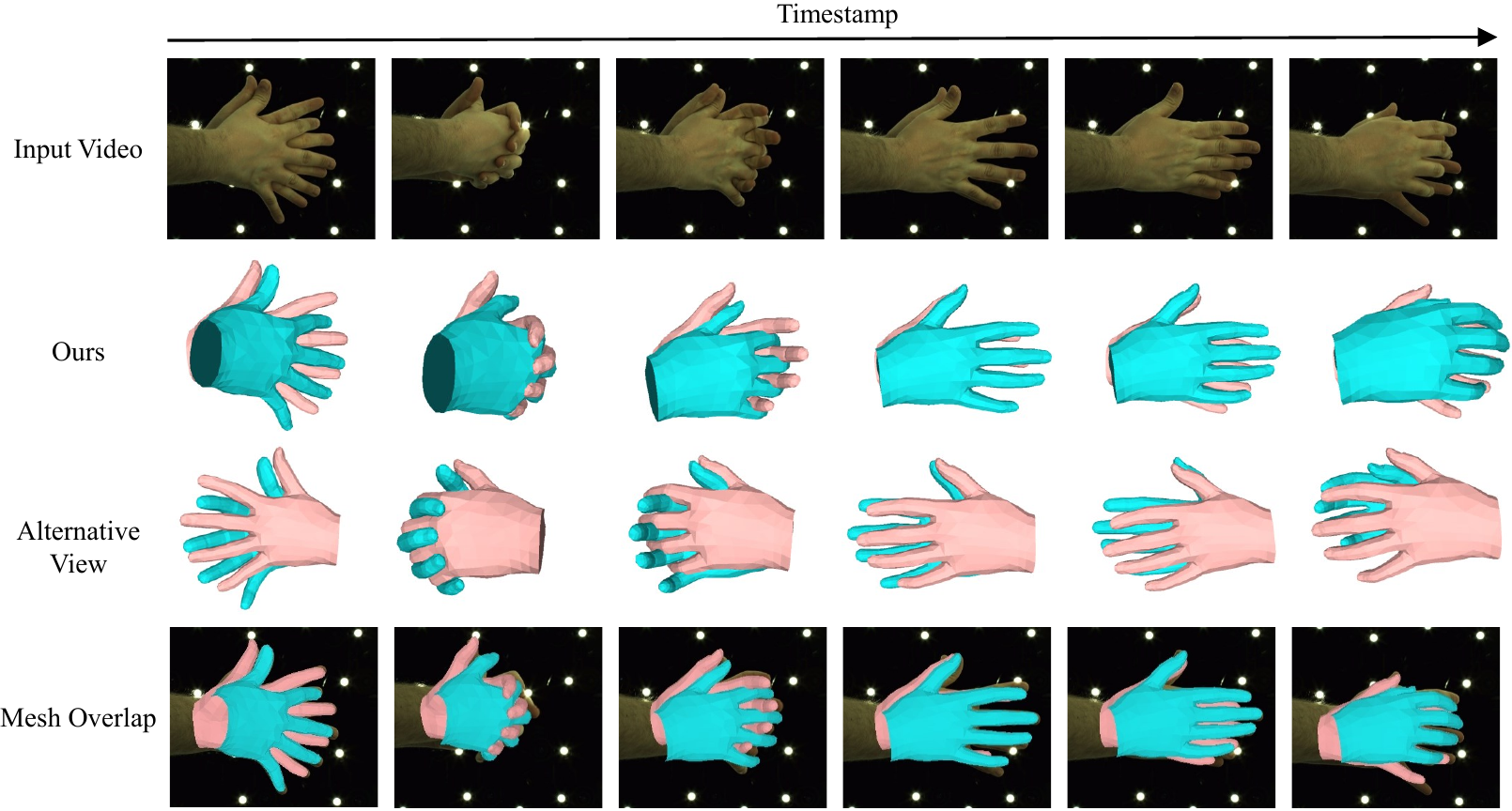}
		\vspace{-1.5em}
	\caption{Qualitative results of interacting hand reconstruction of our proposed method on the InterHand2.6M~\cite{moon2020interhand2} dataset.
		For clarification, we visualize the mesh with the two views.
		Our method achieves high-quality reconstruction performance under different levels of occlusion.}
	\label{fig:InterHand}
	\vspace{-1.5em}
\end{figure*}

\textbf{Acceleration errors.} As shown in Fig.~\ref{accel}, we visualize the acceleration errors of different methods in a selected video clip from the test set of  InterHand2.6M~\cite{moon2020interhand2}. 
The previous methods, \textit{i.e.,} \cite{moon2020interhand2,zhang2021interacting,boukhayma20193d} are based on reconstructing interacting hands from a single RGB image.
However, in cases when self-occlusion and mutual-occlusion occur, these methods easily exhibit significant discrepancies between the predicted outcomes and the ground truth due to limited information. As a result, these methods suffer from the acceleration error spike among the continuous frames once they encounter the aforementioned situation. In our proposed method, our model naturally captures the temporal contexts for interacting hand motion smoothness. Specifically, we design a temporal smooth loss to explicitly constrain hand pose sequence continuously and smoothly. Based on these designs, our model can effectively constrain the coherence of hand movements and predict more accurate interacting hand poses with contextual information under severe occlusions, thus largely solving the appearance of acceleration error spikes. Compared with the temporal method SeqHand~\cite{yang2020seqhand}, our proposed method shows lower acceleration errors along the time step, which indicates temporally consistent 3D
motion output.

\textbf{Interacting hand reconstruction.} In Fig.~\ref{fig:comp_with_others}, we select a video slice from the test set for better displaying the reconstruction results of different methods. As the occlusion of both hands gradually increases, HPS~\cite{boukhayma20193d} and ITH-3D~\cite{zhang2021interacting} generate unreasonable results, \textit{i.e.,} interpenetration between interacting hands, incorrect hand poses. Compared with them, our method captures
more hand details and generates more accurate interacting hand poses.

\begin{figure*}[t]
	\centering
	\includegraphics[width=0.8\textwidth]{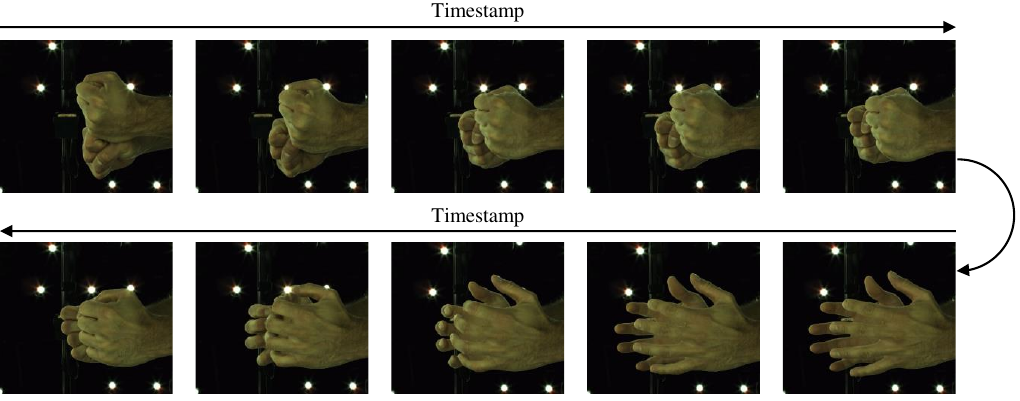}
	\caption{The visualization of input interacting hand sequence.}
	\label{fig:img_visiual}
	 \vspace{-1.0em}
\end{figure*}

\begin{figure*}[t]
	\centering
	\includegraphics[width=0.8\textwidth]{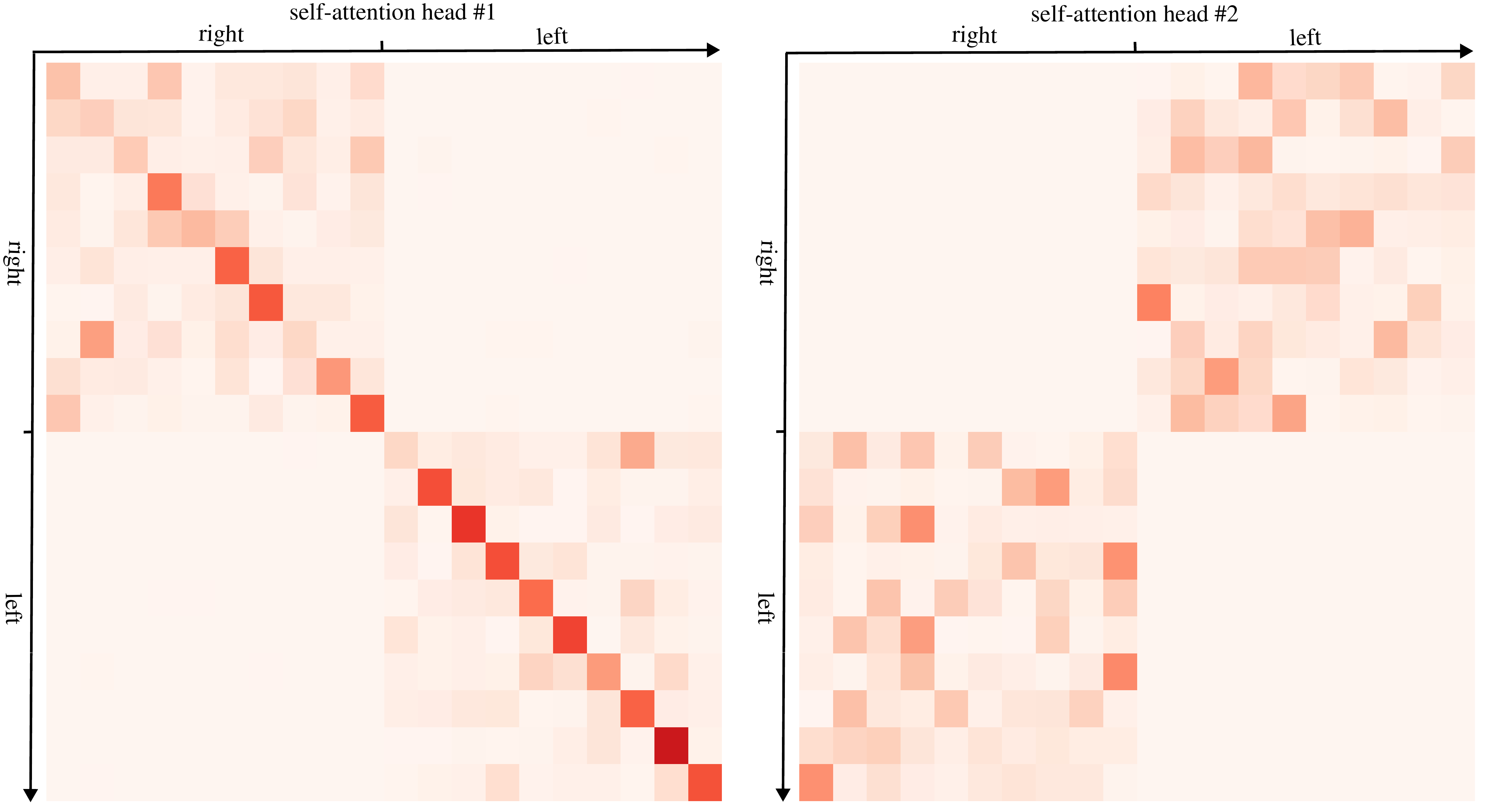}
	\caption{The visualization of attention maps in the multi-head self-attention~(MHSA) block. The darker the color, the higher the corresponding weight.}
	\label{fig:self-attn}
	 \vspace{-1.5em}
\end{figure*}

\textbf{Multi-view visualization.} Mutual-occlusion is a common phenomenon among interacting hands. Thus, we show the reconstruction results of interacting hands under mutual-occlusion from different viewpoints on HIC~\cite{hasson2019learning} and InterHand2.6M~\cite{moon2020interhand2} in Fig.~\ref{fig:HIC} and Fig.~\ref{fig:InterHand}, respectively. Our proposed method could aggregate the temporal information to predict the occluded hand and simultaneously utilize the spatial information to alleviate the collision between both hands. The alternative view of reconstruction results can validate the effectiveness of our proposed method.

\begin{figure*}[t]
	\centering
	\includegraphics[width=0.9\textwidth]{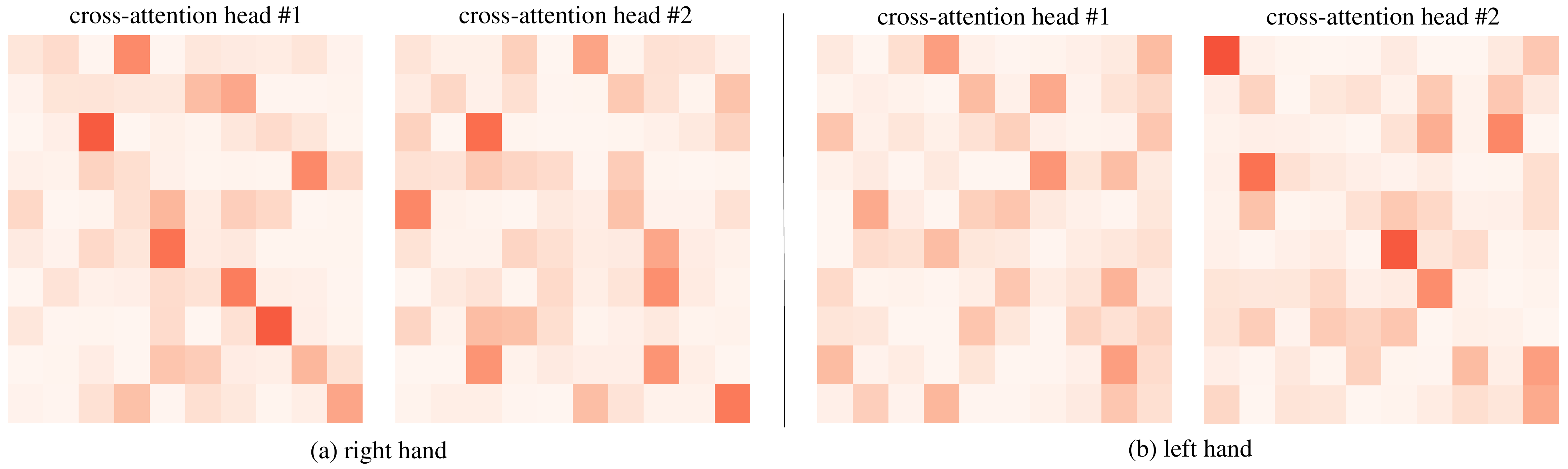}
	\caption{The visualization of attention maps in the multi-head cross-attention~(MHCA) block. The darker the color, the higher the corresponding weight.}
	\label{fig:cross-attn}
\end{figure*}

\textbf{Attention mechanism visualization.}  We visualize the attention mechanisms in the MHSA and MHCA blocks to show how they work. The input video sequence is shown in Fig.~\ref{fig:img_visiual}. As shown in Fig.~\ref{fig:self-attn}, we visualize the first two attention maps of the MHSA block in the first temporal block. In MHSA, we concat both hand feature sequence together. It is observed that different heads could capture effective information from different regions. For instance, the first head in the MHSA block aims to transfer information among each single hand feature sequence. The second head could extract useful information from the feature sequence of the other hand. As shown in Fig.~\ref{fig:cross-attn}, we visualize the first two attention maps of both MHCA blocks in the first temporal block. Both hand feature sequences could mine effective temporal cues from the global feature sequence.

\section{Future Work}
Our proposed method designs an interpenetration detector to detect the predicted interacting hand vertices that are inside each other. This operation needs to calculate the distances from all vertices of both hands, which requires more computational resources. We will explore how to detect collisions more effectively by introducing prior information about collision-prone hand regions. Moreover, we will try to utilize more available weakly labeled data to enhance the generalization and characterization capabilities of our proposed method in the future work.

\section{Conclusion}
In this work, we are dedicated to explicitly exploiting the temporal-spatial information to better reconstruct interacting hands with monocular RGB video. To this end, we we have designed a framework that incorporates a temporal encoder with multi-scale temporal information to capture the temporal context of hand motion, and a smooth objective to enforce temporal consistency.
Additionally, we explore the spatial information by reducing the likelihood of interpenetration between the interacting hands, resulting in more realistic hand representations that adhere to physical constraints. 
Our experiments demonstrate the superiority of our proposed framework over existing state-of-the-art methods.

\section{Acknowledgement}
This work was supported by the GPU cluster built by MCC
Lab of Information Science and Technology Institution and the Supercomputing Center of the
USTC.

\bibliographystyle{ACM-Reference-Format}
\bibliography{sample-base}


\begin{thebibliography}{58}


\ifx \showCODEN    \undefined \def \showCODEN     #1{\unskip}     \fi
\ifx \showDOI      \undefined \def \showDOI       #1{#1}\fi
\ifx \showISBNx    \undefined \def \showISBNx     #1{\unskip}     \fi
\ifx \showISBNxiii \undefined \def \showISBNxiii  #1{\unskip}     \fi
\ifx \showISSN     \undefined \def \showISSN      #1{\unskip}     \fi
\ifx \showLCCN     \undefined \def \showLCCN      #1{\unskip}     \fi
\ifx \shownote     \undefined \def \shownote      #1{#1}          \fi
\ifx \showarticletitle \undefined \def \showarticletitle #1{#1}   \fi
\ifx \showURL      \undefined \def \showURL       {\relax}        \fi
\providecommand\bibfield[2]{#2}
\providecommand\bibinfo[2]{#2}
\providecommand\natexlab[1]{#1}
\providecommand\showeprint[2][]{arXiv:#2}

\bibitem[Ballan et~al\mbox{.}(2012)]%
        {ballan2012motion}
\bibfield{author}{\bibinfo{person}{Luca Ballan}, \bibinfo{person}{Aparna
  Taneja}, \bibinfo{person}{J{\"u}rgen Gall}, \bibinfo{person}{Luc Van~Gool},
  {and} \bibinfo{person}{Marc Pollefeys}.} \bibinfo{year}{2012}\natexlab{}.
\newblock \showarticletitle{Motion capture of hands in action using
  discriminative salient points}. In \bibinfo{booktitle}{\emph{Proceedings of
  the European Conference on Computer Vision}}. \bibinfo{pages}{640--653}.
\newblock


\bibitem[Boukhayma et~al\mbox{.}(2019)]%
        {boukhayma20193d}
\bibfield{author}{\bibinfo{person}{Adnane Boukhayma},
  \bibinfo{person}{Rodrigo~de Bem}, {and} \bibinfo{person}{Philip~HS Torr}.}
  \bibinfo{year}{2019}\natexlab{}.
\newblock \showarticletitle{{3D} hand shape and pose from images in the wild}.
  In \bibinfo{booktitle}{\emph{Proceedings of the IEEE Conference on Computer
  Vision and Pattern Recognition}}. \bibinfo{pages}{10843--10852}.
\newblock


\bibitem[Cai et~al\mbox{.}(2018)]%
        {cai2018weakly}
\bibfield{author}{\bibinfo{person}{Yujun Cai}, \bibinfo{person}{Liuhao Ge},
  \bibinfo{person}{Jianfei Cai}, {and} \bibinfo{person}{Junsong Yuan}.}
  \bibinfo{year}{2018}\natexlab{}.
\newblock \showarticletitle{Weakly-supervised {3D} hand pose estimation from
  monocular {RGB} images}. In \bibinfo{booktitle}{\emph{Proceedings of the
  European Conference on Computer Vision}}. \bibinfo{pages}{666--682}.
\newblock


\bibitem[Chen et~al\mbox{.}(2021)]%
        {chen2021model}
\bibfield{author}{\bibinfo{person}{Yujin Chen}, \bibinfo{person}{Zhigang Tu},
  \bibinfo{person}{Di Kang}, \bibinfo{person}{Linchao Bao},
  \bibinfo{person}{Ying Zhang}, \bibinfo{person}{Xuefei Zhe},
  \bibinfo{person}{Ruizhi Chen}, {and} \bibinfo{person}{Junsong Yuan}.}
  \bibinfo{year}{2021}\natexlab{}.
\newblock \showarticletitle{Model-based {3D} hand reconstruction via
  self-supervised learning}. In \bibinfo{booktitle}{\emph{Proceedings of the
  IEEE Conference on Computer Vision and Pattern Recognition}}.
  \bibinfo{pages}{10451--10460}.
\newblock


\bibitem[Deng et~al\mbox{.}(2009)]%
        {deng2009imagenet}
\bibfield{author}{\bibinfo{person}{Jia Deng}, \bibinfo{person}{Wei Dong},
  \bibinfo{person}{Richard Socher}, \bibinfo{person}{Li-Jia Li},
  \bibinfo{person}{Kai Li}, {and} \bibinfo{person}{Li Fei-Fei}.}
  \bibinfo{year}{2009}\natexlab{}.
\newblock \showarticletitle{Imagenet: A large-scale hierarchical image
  database}. In \bibinfo{booktitle}{\emph{Proceedings of the IEEE Conference on
  Computer Vision and Pattern Recognition}}. \bibinfo{pages}{248--255}.
\newblock


\bibitem[Diffenderfer et~al\mbox{.}(2021)]%
        {diffenderfer2021winning}
\bibfield{author}{\bibinfo{person}{James Diffenderfer}, \bibinfo{person}{Brian
  Bartoldson}, \bibinfo{person}{Shreya Chaganti}, \bibinfo{person}{Jize Zhang},
  {and} \bibinfo{person}{Bhavya Kailkhura}.} \bibinfo{year}{2021}\natexlab{}.
\newblock \showarticletitle{A winning hand: Compressing deep networks can
  improve out-of-distribution robustness}. In
  \bibinfo{booktitle}{\emph{Proceedings of the Advances in Neural Information
  Processing Systems}}, Vol.~\bibinfo{volume}{34}.
\newblock


\bibitem[Fan et~al\mbox{.}(2021b)]%
        {fan2021understanding}
\bibfield{author}{\bibinfo{person}{Hehe Fan}, \bibinfo{person}{Tao Zhuo},
  \bibinfo{person}{Xin Yu}, \bibinfo{person}{Yi Yang}, {and}
  \bibinfo{person}{Mohan Kankanhalli}.} \bibinfo{year}{2021}\natexlab{b}.
\newblock \showarticletitle{Understanding atomic hand-object interaction with
  human intention}.
\newblock \bibinfo{journal}{\emph{IEEE Transactions on Circuits and Systems for
  Video Technology}} \bibinfo{volume}{32}, \bibinfo{number}{1}
  (\bibinfo{year}{2021}), \bibinfo{pages}{275--285}.
\newblock


\bibitem[Fan et~al\mbox{.}(2021a)]%
        {fan2021learning}
\bibfield{author}{\bibinfo{person}{Zicong Fan}, \bibinfo{person}{Adrian Spurr},
  \bibinfo{person}{Muhammed Kocabas}, \bibinfo{person}{Siyu Tang},
  \bibinfo{person}{Michael~J Black}, {and} \bibinfo{person}{Otmar Hilliges}.}
  \bibinfo{year}{2021}\natexlab{a}.
\newblock \showarticletitle{Learning to disambiguate strongly interacting hands
  via probabilistic per-pixel part segmentation}. In
  \bibinfo{booktitle}{\emph{2021 International Conference on 3D Vision}}.
  \bibinfo{pages}{1--10}.
\newblock


\bibitem[Grady et~al\mbox{.}(2021)]%
        {grady2021contactopt}
\bibfield{author}{\bibinfo{person}{Patrick Grady}, \bibinfo{person}{Chengcheng
  Tang}, \bibinfo{person}{Christopher~D Twigg}, \bibinfo{person}{Minh Vo},
  \bibinfo{person}{Samarth Brahmbhatt}, {and} \bibinfo{person}{Charles~C
  Kemp}.} \bibinfo{year}{2021}\natexlab{}.
\newblock \showarticletitle{Contactopt: Optimizing contact to improve grasps}.
  In \bibinfo{booktitle}{\emph{Proceedings of the IEEE Conference on Computer
  Vision and Pattern Recognition}}. \bibinfo{pages}{1471--1481}.
\newblock


\bibitem[Guo et~al\mbox{.}(2017)]%
        {guo2017region}
\bibfield{author}{\bibinfo{person}{Hengkai Guo}, \bibinfo{person}{Guijin Wang},
  \bibinfo{person}{Xinghao Chen}, \bibinfo{person}{Cairong Zhang},
  \bibinfo{person}{Fei Qiao}, {and} \bibinfo{person}{Huazhong Yang}.}
  \bibinfo{year}{2017}\natexlab{}.
\newblock \showarticletitle{Region ensemble network: {Improving} convolutional
  network for hand pose estimation}. In \bibinfo{booktitle}{\emph{Proceedings
  of the IEEE International Conference on Image Processing}}.
  \bibinfo{pages}{4512--4516}.
\newblock


\bibitem[Han et~al\mbox{.}(2020)]%
        {han2020megatrack}
\bibfield{author}{\bibinfo{person}{Shangchen Han}, \bibinfo{person}{Beibei
  Liu}, \bibinfo{person}{Randi Cabezas}, \bibinfo{person}{Christopher~D Twigg},
  \bibinfo{person}{Peizhao Zhang}, \bibinfo{person}{Jeff Petkau},
  \bibinfo{person}{Tsz-Ho Yu}, \bibinfo{person}{Chun-Jung Tai},
  \bibinfo{person}{Muzaffer Akbay}, \bibinfo{person}{Zheng Wang},
  {et~al\mbox{.}}} \bibinfo{year}{2020}\natexlab{}.
\newblock \showarticletitle{{MEgATrack}: monochrome egocentric articulated
  hand-tracking for virtual reality.}
\newblock \bibinfo{journal}{\emph{ACM Transactions on Graphics}}
  \bibinfo{volume}{39}, \bibinfo{number}{4} (\bibinfo{year}{2020}),
  \bibinfo{pages}{87}.
\newblock


\bibitem[Hasson et~al\mbox{.}(2020)]%
        {hasson2020leveraging}
\bibfield{author}{\bibinfo{person}{Yana Hasson}, \bibinfo{person}{Bugra Tekin},
  \bibinfo{person}{Federica Bogo}, \bibinfo{person}{Ivan Laptev},
  \bibinfo{person}{Marc Pollefeys}, {and} \bibinfo{person}{Cordelia Schmid}.}
  \bibinfo{year}{2020}\natexlab{}.
\newblock \showarticletitle{Leveraging photometric consistency over time for
  sparsely supervised hand-object reconstruction}. In
  \bibinfo{booktitle}{\emph{Proceedings of the IEEE Conference on Computer
  Vision and Pattern Recognition}}. \bibinfo{pages}{571--580}.
\newblock


\bibitem[Hasson et~al\mbox{.}(2019)]%
        {hasson2019learning}
\bibfield{author}{\bibinfo{person}{Yana Hasson}, \bibinfo{person}{Gul Varol},
  \bibinfo{person}{Dimitrios Tzionas}, \bibinfo{person}{Igor Kalevatykh},
  \bibinfo{person}{Michael~J Black}, \bibinfo{person}{Ivan Laptev}, {and}
  \bibinfo{person}{Cordelia Schmid}.} \bibinfo{year}{2019}\natexlab{}.
\newblock \showarticletitle{Learning joint reconstruction of hands and
  manipulated objects}. In \bibinfo{booktitle}{\emph{Proceedings of the IEEE
  Conference on Computer Vision and Pattern Recognition}}.
  \bibinfo{pages}{11807--11816}.
\newblock


\bibitem[He et~al\mbox{.}(2016)]%
        {he2016deep}
\bibfield{author}{\bibinfo{person}{Kaiming He}, \bibinfo{person}{Xiangyu
  Zhang}, \bibinfo{person}{Shaoqing Ren}, {and} \bibinfo{person}{Jian Sun}.}
  \bibinfo{year}{2016}\natexlab{}.
\newblock \showarticletitle{Deep residual learning for image recognition}. In
  \bibinfo{booktitle}{\emph{Proceedings of the IEEE Conference on Computer
  Vision and Pattern Recognition}}. \bibinfo{pages}{770--778}.
\newblock


\bibitem[Hu et~al\mbox{.}(2023)]%
        {hu2023signbert+}
\bibfield{author}{\bibinfo{person}{Hezhen Hu}, \bibinfo{person}{Weichao Zhao},
  \bibinfo{person}{Wengang Zhou}, {and} \bibinfo{person}{Houqiang Li}.}
  \bibinfo{year}{2023}\natexlab{}.
\newblock \showarticletitle{SignBERT+: Hand-model-aware Self-supervised
  Pre-training for Sign Language Understanding}.
\newblock \bibinfo{journal}{\emph{IEEE Transactions on Pattern Analysis and
  Machine Intelligence}} (\bibinfo{year}{2023}).
\newblock


\bibitem[Hu et~al\mbox{.}(2021)]%
        {hu2021hand}
\bibfield{author}{\bibinfo{person}{Hezhen Hu}, \bibinfo{person}{Wengang Zhou},
  {and} \bibinfo{person}{Houqiang Li}.} \bibinfo{year}{2021}\natexlab{}.
\newblock \showarticletitle{Hand-Model-Aware Sign Language Recognition}. In
  \bibinfo{booktitle}{\emph{Proceedings of the AAAI Conference on Artificial
  Intelligence}}. \bibinfo{pages}{1558--1566}.
\newblock


\bibitem[Huang et~al\mbox{.}(2017)]%
        {huang2017ego}
\bibfield{author}{\bibinfo{person}{Shao Huang}, \bibinfo{person}{Weiqiang
  Wang}, \bibinfo{person}{Shengfeng He}, {and} \bibinfo{person}{Rynson W.~H.
  Lau}.} \bibinfo{year}{2017}\natexlab{}.
\newblock \showarticletitle{Egocentric Hand Detection Via Dynamic Region
  Growing}.
\newblock \bibinfo{journal}{\emph{ACM Transaction on Multimedia Computing,
  Communication, and Applications}} \bibinfo{volume}{14}, \bibinfo{number}{1}
  (\bibinfo{year}{2017}), \bibinfo{pages}{1--17}.
\newblock


\bibitem[Huang et~al\mbox{.}(2020)]%
        {huang2020awr}
\bibfield{author}{\bibinfo{person}{Weiting Huang}, \bibinfo{person}{Pengfei
  Ren}, \bibinfo{person}{Jingyu Wang}, \bibinfo{person}{Qi Qi}, {and}
  \bibinfo{person}{Haifeng Sun}.} \bibinfo{year}{2020}\natexlab{}.
\newblock \showarticletitle{{AWR}: Adaptive weighting regression for {3D} hand
  pose estimation}. In \bibinfo{booktitle}{\emph{Proceedings of the AAAI
  Conference on Artificial Intelligence}}. \bibinfo{pages}{11061--11068}.
\newblock


\bibitem[Iqbal et~al\mbox{.}(2018)]%
        {iqbal2018hand}
\bibfield{author}{\bibinfo{person}{Umar Iqbal}, \bibinfo{person}{Pavlo
  Molchanov}, \bibinfo{person}{Thomas Breuel~Juergen Gall}, {and}
  \bibinfo{person}{Jan Kautz}.} \bibinfo{year}{2018}\natexlab{}.
\newblock \showarticletitle{Hand pose estimation via latent 2.5{D} heatmap
  regression}. In \bibinfo{booktitle}{\emph{Proceedings of the European
  Conference on Computer Vision}}. \bibinfo{pages}{118--134}.
\newblock


\bibitem[Jiao et~al\mbox{.}(2022)]%
        {jiaoglpose}
\bibfield{author}{\bibinfo{person}{Yingying Jiao}, \bibinfo{person}{Haipeng
  Chen}, \bibinfo{person}{Runyang Feng}, \bibinfo{person}{Haoming Chen},
  \bibinfo{person}{Sifan Wu}, \bibinfo{person}{Yifang Yin}, {and}
  \bibinfo{person}{Zhenguang Liu}.} \bibinfo{year}{2022}\natexlab{}.
\newblock \showarticletitle{{GLPose}: Global-{Local} Representation Learning
  for Human Pose Estimation}.
\newblock \bibinfo{journal}{\emph{ACM Transaction on Multimedia Computing,
  Communication, and Applications}} \bibinfo{volume}{18}, \bibinfo{number}{25}
  (\bibinfo{year}{2022}), \bibinfo{pages}{1--16}.
\newblock


\bibitem[Kanazawa et~al\mbox{.}(2019)]%
        {kanazawa2019learning}
\bibfield{author}{\bibinfo{person}{Angjoo Kanazawa}, \bibinfo{person}{Jason~Y
  Zhang}, \bibinfo{person}{Panna Felsen}, {and} \bibinfo{person}{Jitendra
  Malik}.} \bibinfo{year}{2019}\natexlab{}.
\newblock \showarticletitle{Learning {3D} human dynamics from video}. In
  \bibinfo{booktitle}{\emph{Proceedings of the IEEE Conference on Computer
  Vision and Pattern Recognition}}. \bibinfo{pages}{5614--5623}.
\newblock


\bibitem[Kim et~al\mbox{.}(2021)]%
        {kim2021end}
\bibfield{author}{\bibinfo{person}{Dong~Uk Kim}, \bibinfo{person}{Kwang~In
  Kim}, {and} \bibinfo{person}{Seungryul Baek}.}
  \bibinfo{year}{2021}\natexlab{}.
\newblock \showarticletitle{End-to-end detection and pose estimation of two
  interacting hands}. In \bibinfo{booktitle}{\emph{Proceedings of the IEEE
  International Conference on Computer Vision}}. \bibinfo{pages}{11189--11198}.
\newblock


\bibitem[Kingma and Ba(2014)]%
        {kingma2014adam}
\bibfield{author}{\bibinfo{person}{Diederik~P Kingma} {and}
  \bibinfo{person}{Jimmy Ba}.} \bibinfo{year}{2014}\natexlab{}.
\newblock \showarticletitle{Adam: A method for stochastic optimization}.
\newblock \bibinfo{journal}{\emph{arXiv preprint arXiv:1412.6980}}
  (\bibinfo{year}{2014}).
\newblock


\bibitem[Li et~al\mbox{.}(2022)]%
        {Li2022intaghand}
\bibfield{author}{\bibinfo{person}{Mengcheng Li}, \bibinfo{person}{Liang An},
  \bibinfo{person}{Hongwen Zhang}, \bibinfo{person}{Lianpeng Wu},
  \bibinfo{person}{Feng Chen}, \bibinfo{person}{Tao Yu}, {and}
  \bibinfo{person}{Yebin Liu}.} \bibinfo{year}{2022}\natexlab{}.
\newblock \showarticletitle{Interacting Attention Graph for Single Image
  Two-Hand Reconstruction}. In \bibinfo{booktitle}{\emph{Proceedings of the
  IEEE Conference on Computer Vision and Pattern Recognition}}.
  \bibinfo{pages}{2761--2770}.
\newblock


\bibitem[Li et~al\mbox{.}(2020)]%
        {li2020exploiting}
\bibfield{author}{\bibinfo{person}{Moran Li}, \bibinfo{person}{Yuan Gao}, {and}
  \bibinfo{person}{Nong Sang}.} \bibinfo{year}{2020}\natexlab{}.
\newblock \showarticletitle{Exploiting learnable joint groups for hand pose
  estimation}.
\newblock \bibinfo{journal}{\emph{arXiv preprint arXiv:2012.09496}}
  (\bibinfo{year}{2020}).
\newblock


\bibitem[Liang et~al\mbox{.}(2014)]%
        {liang2014resolving}
\bibfield{author}{\bibinfo{person}{Hui Liang}, \bibinfo{person}{Junsong Yuan},
  {and} \bibinfo{person}{Daniel Thalmann}.} \bibinfo{year}{2014}\natexlab{}.
\newblock \showarticletitle{Resolving ambiguous hand pose predictions by
  exploiting part correlations}.
\newblock \bibinfo{journal}{\emph{IEEE Transactions on Circuits and Systems for
  Video Technology}} \bibinfo{volume}{25}, \bibinfo{number}{7}
  (\bibinfo{year}{2014}), \bibinfo{pages}{1125--1139}.
\newblock


\bibitem[Lin et~al\mbox{.}(2021)]%
        {lin2021two}
\bibfield{author}{\bibinfo{person}{Fanqing Lin}, \bibinfo{person}{Connor
  Wilhelm}, {and} \bibinfo{person}{Tony Martinez}.}
  \bibinfo{year}{2021}\natexlab{}.
\newblock \showarticletitle{Two-hand global {3D} pose estimation using
  monocular {RGB}}. In \bibinfo{booktitle}{\emph{Proceedings of the IEEE Winter
  Conference on Applications of Computer Vision}}. \bibinfo{pages}{2373--2381}.
\newblock


\bibitem[Liu et~al\mbox{.}(2021)]%
        {liu2021han}
\bibfield{author}{\bibinfo{person}{Jianbo Liu}, \bibinfo{person}{Ying Wang},
  \bibinfo{person}{Shiming Xiang}, {and} \bibinfo{person}{Chunhong Pan}.}
  \bibinfo{year}{2021}\natexlab{}.
\newblock \showarticletitle{{HAN}: An efficient hierarchical self-attention
  network for skeleton-based gesture recognition}.
\newblock \bibinfo{journal}{\emph{arXiv preprint arXiv:2106.13391}}
  (\bibinfo{year}{2021}).
\newblock


\bibitem[Meng et~al\mbox{.}(2022)]%
        {meng20223d}
\bibfield{author}{\bibinfo{person}{Hao Meng}, \bibinfo{person}{Sheng Jin},
  \bibinfo{person}{Wentao Liu}, \bibinfo{person}{Chen Qian},
  \bibinfo{person}{Mengxiang Lin}, \bibinfo{person}{Wanli Ouyang}, {and}
  \bibinfo{person}{Ping Luo}.} \bibinfo{year}{2022}\natexlab{}.
\newblock \showarticletitle{{3D} interacting hand pose estimation by hand
  de-occlusion and removal}. In \bibinfo{booktitle}{\emph{Proceedings of the
  European Conference on Computer Vision}}. \bibinfo{pages}{380--397}.
\newblock


\bibitem[Moon et~al\mbox{.}(2018)]%
        {moon2018v2v}
\bibfield{author}{\bibinfo{person}{Gyeongsik Moon}, \bibinfo{person}{Ju~Yong
  Chang}, {and} \bibinfo{person}{Kyoung~Mu Lee}.}
  \bibinfo{year}{2018}\natexlab{}.
\newblock \showarticletitle{V2v-posenet: {Voxel}-to-voxel prediction network
  for accurate {3D} hand and human pose estimation from a single depth map}. In
  \bibinfo{booktitle}{\emph{Proceedings of the IEEE Conference on Computer
  Vision and Pattern Recognition}}. \bibinfo{pages}{5079--5088}.
\newblock


\bibitem[Moon et~al\mbox{.}(2020)]%
        {moon2020interhand2}
\bibfield{author}{\bibinfo{person}{Gyeongsik Moon}, \bibinfo{person}{Shoou-I
  Yu}, \bibinfo{person}{He Wen}, \bibinfo{person}{Takaaki Shiratori}, {and}
  \bibinfo{person}{Kyoung~Mu Lee}.} \bibinfo{year}{2020}\natexlab{}.
\newblock \showarticletitle{InterHand2. 6M: {A} dataset and baseline for {3D}
  interacting hand pose estimation from a single {RGB} image}. In
  \bibinfo{booktitle}{\emph{Proceedings of the European Conference on Computer
  Vision}}. \bibinfo{pages}{548--564}.
\newblock


\bibitem[Mueller et~al\mbox{.}(2019)]%
        {mueller2019real}
\bibfield{author}{\bibinfo{person}{Franziska Mueller}, \bibinfo{person}{Micah
  Davis}, \bibinfo{person}{Florian Bernard}, \bibinfo{person}{Oleksandr
  Sotnychenko}, \bibinfo{person}{Mickeal Verschoor}, \bibinfo{person}{Miguel~A
  Otaduy}, \bibinfo{person}{Dan Casas}, {and} \bibinfo{person}{Christian
  Theobalt}.} \bibinfo{year}{2019}\natexlab{}.
\newblock \showarticletitle{Real-time pose and shape reconstruction of two
  interacting hands with a single depth camera}.
\newblock \bibinfo{journal}{\emph{ACM Transactions on Graphics}}
  \bibinfo{volume}{38}, \bibinfo{number}{4} (\bibinfo{year}{2019}),
  \bibinfo{pages}{1--13}.
\newblock


\bibitem[Narasimhaswamy et~al\mbox{.}(2020)]%
        {narasimhaswamy2020detecting}
\bibfield{author}{\bibinfo{person}{Supreeth Narasimhaswamy},
  \bibinfo{person}{Trung Nguyen}, {and} \bibinfo{person}{Minh~Hoai Nguyen}.}
  \bibinfo{year}{2020}\natexlab{}.
\newblock \showarticletitle{Detecting hands and recognizing physical contact in
  the wild}. In \bibinfo{booktitle}{\emph{Proceedings of the Advances in Neural
  Information Processing Systems}}, Vol.~\bibinfo{volume}{33}.
  \bibinfo{pages}{7841--7851}.
\newblock


\bibitem[Paszke et~al\mbox{.}(2019)]%
        {paszke2019pytorch}
\bibfield{author}{\bibinfo{person}{Adam Paszke}, \bibinfo{person}{Sam Gross},
  \bibinfo{person}{Francisco Massa}, \bibinfo{person}{Adam Lerer},
  \bibinfo{person}{James Bradbury}, \bibinfo{person}{Gregory Chanan},
  \bibinfo{person}{Trevor Killeen}, \bibinfo{person}{Zeming Lin},
  \bibinfo{person}{Natalia Gimelshein}, \bibinfo{person}{Luca Antiga},
  {et~al\mbox{.}}} \bibinfo{year}{2019}\natexlab{}.
\newblock \showarticletitle{Pytorch: {An} imperative style, high-performance
  deep learning library}. In \bibinfo{booktitle}{\emph{Proceedings of the
  Advances in Neural Information Processing Systems}},
  Vol.~\bibinfo{volume}{32}. \bibinfo{pages}{8026--8037}.
\newblock


\bibitem[Qian et~al\mbox{.}(2014)]%
        {qian2014realtime}
\bibfield{author}{\bibinfo{person}{Chen Qian}, \bibinfo{person}{Xiao Sun},
  \bibinfo{person}{Yichen Wei}, \bibinfo{person}{Xiaoou Tang}, {and}
  \bibinfo{person}{Jian Sun}.} \bibinfo{year}{2014}\natexlab{}.
\newblock \showarticletitle{Realtime and robust hand tracking from depth}. In
  \bibinfo{booktitle}{\emph{Proceedings of the IEEE Conference on Computer
  Vision and Pattern Recognition}}. \bibinfo{pages}{1106--1113}.
\newblock


\bibitem[Romero et~al\mbox{.}(2010)]%
        {romero2010hands}
\bibfield{author}{\bibinfo{person}{Javier Romero}, \bibinfo{person}{Hedvig
  Kjellstr{\"o}m}, {and} \bibinfo{person}{Danica Kragic}.}
  \bibinfo{year}{2010}\natexlab{}.
\newblock \showarticletitle{Hands in action: real-time {3D} reconstruction of
  hands in interaction with objects}. In \bibinfo{booktitle}{\emph{Proceedings
  of IEEE International Conference on Robotics and Automation}}.
  \bibinfo{pages}{458--463}.
\newblock


\bibitem[Romero et~al\mbox{.}(2017)]%
        {romero2017embodied}
\bibfield{author}{\bibinfo{person}{Javier Romero}, \bibinfo{person}{Dimitrios
  Tzionas}, {and} \bibinfo{person}{Michael~J Black}.}
  \bibinfo{year}{2017}\natexlab{}.
\newblock \showarticletitle{Embodied hands: {Modeling} and capturing hands and
  bodies together}.
\newblock \bibinfo{journal}{\emph{ACM Transactions on Graphics}}
  \bibinfo{volume}{36}, \bibinfo{number}{6} (\bibinfo{year}{2017}),
  \bibinfo{pages}{1--17}.
\newblock


\bibitem[Sanchez-Riera et~al\mbox{.}(2017)]%
        {sanchez2017robust}
\bibfield{author}{\bibinfo{person}{Jordi Sanchez-Riera},
  \bibinfo{person}{Kathiravan Srinivasan}, \bibinfo{person}{Kai-Lung Hua},
  \bibinfo{person}{Wen-Huang Cheng}, \bibinfo{person}{M~Anwar Hossain}, {and}
  \bibinfo{person}{Mohammed~F Alhamid}.} \bibinfo{year}{2017}\natexlab{}.
\newblock \showarticletitle{Robust {RGB-D} hand tracking using deep learning
  priors}.
\newblock \bibinfo{journal}{\emph{IEEE Transactions on Circuits and Systems for
  Video Technology}} \bibinfo{volume}{28}, \bibinfo{number}{9}
  (\bibinfo{year}{2017}), \bibinfo{pages}{2289--2301}.
\newblock


\bibitem[Shan et~al\mbox{.}(2021)]%
        {shan2021cohesiv}
\bibfield{author}{\bibinfo{person}{Dandan Shan}, \bibinfo{person}{Richard
  Higgins}, {and} \bibinfo{person}{David Fouhey}.}
  \bibinfo{year}{2021}\natexlab{}.
\newblock \showarticletitle{COHESIV: Contrastive Object and Hand Embedding
  Segmentation In Video}. In \bibinfo{booktitle}{\emph{Proceedings of the
  Advances in Neural Information Processing Systems}},
  Vol.~\bibinfo{volume}{34}. \bibinfo{pages}{5898--5909}.
\newblock


\bibitem[Sharp et~al\mbox{.}(2015)]%
        {sharp2015accurate}
\bibfield{author}{\bibinfo{person}{Toby Sharp}, \bibinfo{person}{Cem Keskin},
  \bibinfo{person}{Duncan Robertson}, \bibinfo{person}{Jonathan Taylor},
  \bibinfo{person}{Jamie Shotton}, \bibinfo{person}{David Kim},
  \bibinfo{person}{Christoph Rhemann}, \bibinfo{person}{Ido Leichter},
  \bibinfo{person}{Alon Vinnikov}, \bibinfo{person}{Yichen Wei},
  {et~al\mbox{.}}} \bibinfo{year}{2015}\natexlab{}.
\newblock \showarticletitle{Accurate, robust, and flexible real-time hand
  tracking}. In \bibinfo{booktitle}{\emph{Proceedings of the 33rd Annual ACM
  Conference on Human Factors in Computing Systems}}.
  \bibinfo{pages}{3633--3642}.
\newblock


\bibitem[Smith et~al\mbox{.}(2020)]%
        {smith2020constraining}
\bibfield{author}{\bibinfo{person}{Breannan Smith}, \bibinfo{person}{Chenglei
  Wu}, \bibinfo{person}{He Wen}, \bibinfo{person}{Patrick Peluse},
  \bibinfo{person}{Yaser Sheikh}, \bibinfo{person}{Jessica~K Hodgins}, {and}
  \bibinfo{person}{Takaaki Shiratori}.} \bibinfo{year}{2020}\natexlab{}.
\newblock \showarticletitle{Constraining dense hand surface tracking with
  elasticity}.
\newblock \bibinfo{journal}{\emph{ACM Transactions on Graphics}}
  \bibinfo{volume}{39}, \bibinfo{number}{6} (\bibinfo{year}{2020}),
  \bibinfo{pages}{1--14}.
\newblock


\bibitem[Tagliasacchi et~al\mbox{.}(2015)]%
        {tagliasacchi2015robust}
\bibfield{author}{\bibinfo{person}{Andrea Tagliasacchi},
  \bibinfo{person}{Matthias Schr{\"o}der}, \bibinfo{person}{Anastasia Tkach},
  \bibinfo{person}{Sofien Bouaziz}, \bibinfo{person}{Mario Botsch}, {and}
  \bibinfo{person}{Mark Pauly}.} \bibinfo{year}{2015}\natexlab{}.
\newblock \showarticletitle{Robust articulated-icp for real-time hand
  tracking}. In \bibinfo{booktitle}{\emph{Proceedings of the Computer Graphics
  Forum}}. \bibinfo{pages}{101--114}.
\newblock


\bibitem[Tang et~al\mbox{.}(2015)]%
        {tang2015opening}
\bibfield{author}{\bibinfo{person}{Danhang Tang}, \bibinfo{person}{Jonathan
  Taylor}, \bibinfo{person}{Pushmeet Kohli}, \bibinfo{person}{Cem Keskin},
  \bibinfo{person}{Tae-Kyun Kim}, {and} \bibinfo{person}{Jamie Shotton}.}
  \bibinfo{year}{2015}\natexlab{}.
\newblock \showarticletitle{Opening the black box: Hierarchical sampling
  optimization for estimating human hand pose}. In
  \bibinfo{booktitle}{\emph{Proceedings of the IEEE International Conference on
  Computer Vision}}. \bibinfo{pages}{3325--3333}.
\newblock


\bibitem[Tompson et~al\mbox{.}(2014)]%
        {tompson2014real}
\bibfield{author}{\bibinfo{person}{Jonathan Tompson}, \bibinfo{person}{Murphy
  Stein}, \bibinfo{person}{Yann Lecun}, {and} \bibinfo{person}{Ken Perlin}.}
  \bibinfo{year}{2014}\natexlab{}.
\newblock \showarticletitle{Real-time continuous pose recovery of human hands
  using convolutional networks}.
\newblock \bibinfo{journal}{\emph{ACM Transactions on Graphics}}
  \bibinfo{volume}{33}, \bibinfo{number}{5} (\bibinfo{year}{2014}),
  \bibinfo{pages}{1--10}.
\newblock


\bibitem[Tzionas and Gall(2015)]%
        {tzionas20153d}
\bibfield{author}{\bibinfo{person}{Dimitrios Tzionas} {and}
  \bibinfo{person}{Juergen Gall}.} \bibinfo{year}{2015}\natexlab{}.
\newblock \showarticletitle{{3D} object reconstruction from hand-object
  interactions}. In \bibinfo{booktitle}{\emph{Proceedings of the IEEE
  International Conference on Computer Vision}}. \bibinfo{pages}{729--737}.
\newblock


\bibitem[Vaswani et~al\mbox{.}(2017)]%
        {vaswani2017attention}
\bibfield{author}{\bibinfo{person}{Ashish Vaswani}, \bibinfo{person}{Noam
  Shazeer}, \bibinfo{person}{Niki Parmar}, \bibinfo{person}{Jakob Uszkoreit},
  \bibinfo{person}{Llion Jones}, \bibinfo{person}{Aidan~N Gomez},
  \bibinfo{person}{{\L}ukasz Kaiser}, {and} \bibinfo{person}{Illia
  Polosukhin}.} \bibinfo{year}{2017}\natexlab{}.
\newblock \showarticletitle{Attention is all you need}. In
  \bibinfo{booktitle}{\emph{Proceedings of the Advances in Neural Information
  Processing Systems}}. \bibinfo{pages}{5998--6008}.
\newblock


\bibitem[Wan et~al\mbox{.}(2019)]%
        {wan2019self}
\bibfield{author}{\bibinfo{person}{Chengde Wan}, \bibinfo{person}{Thomas
  Probst}, \bibinfo{person}{Luc~Van Gool}, {and} \bibinfo{person}{Angela Yao}.}
  \bibinfo{year}{2019}\natexlab{}.
\newblock \showarticletitle{Self-supervised {3D} hand pose estimation through
  training by fitting}. In \bibinfo{booktitle}{\emph{Proceedings of the IEEE
  Conference on Computer Vision and Pattern Recognition}}.
  \bibinfo{pages}{10853--10862}.
\newblock


\bibitem[Wang et~al\mbox{.}(2020)]%
        {wang2020rgb2hands}
\bibfield{author}{\bibinfo{person}{Jiayi Wang}, \bibinfo{person}{Franziska
  Mueller}, \bibinfo{person}{Florian Bernard}, \bibinfo{person}{Suzanne Sorli},
  \bibinfo{person}{Oleksandr Sotnychenko}, \bibinfo{person}{Neng Qian},
  \bibinfo{person}{Miguel~A Otaduy}, \bibinfo{person}{Dan Casas}, {and}
  \bibinfo{person}{Christian Theobalt}.} \bibinfo{year}{2020}\natexlab{}.
\newblock \showarticletitle{{RGB2Hands}: real-time tracking of {3D} hand
  interactions from monocular {RGB} video}.
\newblock \bibinfo{journal}{\emph{ACM Transactions on Graphics}}
  \bibinfo{volume}{39}, \bibinfo{number}{6} (\bibinfo{year}{2020}),
  \bibinfo{pages}{1--16}.
\newblock


\bibitem[Wang et~al\mbox{.}(2013)]%
        {wang2013video}
\bibfield{author}{\bibinfo{person}{Yangang Wang}, \bibinfo{person}{Jianyuan
  Min}, \bibinfo{person}{Jianjie Zhang}, \bibinfo{person}{Yebin Liu},
  \bibinfo{person}{Feng Xu}, \bibinfo{person}{Qionghai Dai}, {and}
  \bibinfo{person}{Jinxiang Chai}.} \bibinfo{year}{2013}\natexlab{}.
\newblock \showarticletitle{Video-based hand manipulation capture through
  composite motion control}.
\newblock \bibinfo{journal}{\emph{ACM Transactions on Graphics}}
  \bibinfo{volume}{32}, \bibinfo{number}{4} (\bibinfo{year}{2013}),
  \bibinfo{pages}{1--14}.
\newblock


\bibitem[Wang et~al\mbox{.}(2018)]%
        {wang2018mask}
\bibfield{author}{\bibinfo{person}{Yangang Wang}, \bibinfo{person}{Cong Peng},
  {and} \bibinfo{person}{Yebin Liu}.} \bibinfo{year}{2018}\natexlab{}.
\newblock \showarticletitle{Mask-pose cascaded cnn for {2D} hand pose
  estimation from single color image}.
\newblock \bibinfo{journal}{\emph{IEEE Transactions on Circuits and Systems for
  Video Technology}} \bibinfo{volume}{29}, \bibinfo{number}{11}
  (\bibinfo{year}{2018}), \bibinfo{pages}{3258--3268}.
\newblock


\bibitem[Xiao et~al\mbox{.}(2022)]%
        {xiao22real}
\bibfield{author}{\bibinfo{person}{Yi Xiao}, \bibinfo{person}{Tong Liu},
  \bibinfo{person}{Yu Han}, \bibinfo{person}{Yue Liu}, {and}
  \bibinfo{person}{Yongtian Wang}.} \bibinfo{year}{2022}\natexlab{}.
\newblock \showarticletitle{Realtime Recognition of Dynamic Hand Gestures in
  Practical Applications}.
\newblock \bibinfo{journal}{\emph{ACM Transaction on Multimedia Computing,
  Communication, and Applications}} \bibinfo{volume}{18}, \bibinfo{number}{25}
  (\bibinfo{year}{2022}), \bibinfo{pages}{1--16}.
\newblock


\bibitem[Xu et~al\mbox{.}(2020)]%
        {xu2020improve}
\bibfield{author}{\bibinfo{person}{Lu Xu}, \bibinfo{person}{Chen Hu},
  \bibinfo{person}{Jianru Xue}, \bibinfo{person}{Kuizhi Mei}, {et~al\mbox{.}}}
  \bibinfo{year}{2020}\natexlab{}.
\newblock \showarticletitle{Improve Regression Network on Depth Hand Pose
  Estimation With Auxiliary Variable}.
\newblock \bibinfo{journal}{\emph{IEEE Transactions on Circuits and Systems for
  Video Technology}} \bibinfo{volume}{31}, \bibinfo{number}{3}
  (\bibinfo{year}{2020}), \bibinfo{pages}{890--904}.
\newblock


\bibitem[Yang et~al\mbox{.}(2020)]%
        {yang2020seqhand}
\bibfield{author}{\bibinfo{person}{John Yang}, \bibinfo{person}{Hyung~Jin
  Chang}, \bibinfo{person}{Seungeui Lee}, {and} \bibinfo{person}{Nojun Kwak}.}
  \bibinfo{year}{2020}\natexlab{}.
\newblock \showarticletitle{Seqhand: {RGB}-sequence-based {3D} hand pose and
  shape estimation}. In \bibinfo{booktitle}{\emph{Proceedings of the European
  Conference on Computer Vision}}. \bibinfo{pages}{122--139}.
\newblock


\bibitem[Yang et~al\mbox{.}(2021)]%
        {yang2021cpf}
\bibfield{author}{\bibinfo{person}{Lixin Yang}, \bibinfo{person}{Xinyu Zhan},
  \bibinfo{person}{Kailin Li}, \bibinfo{person}{Wenqiang Xu},
  \bibinfo{person}{Jiefeng Li}, {and} \bibinfo{person}{Cewu Lu}.}
  \bibinfo{year}{2021}\natexlab{}.
\newblock \showarticletitle{CPF: Learning a contact potential field to model
  the hand-object interaction}. In \bibinfo{booktitle}{\emph{Proceedings of the
  IEEE Conference on Computer Vision and Pattern Recognition}}.
  \bibinfo{pages}{11097--11106}.
\newblock


\bibitem[Zhang et~al\mbox{.}(2021)]%
        {zhang2021interacting}
\bibfield{author}{\bibinfo{person}{Baowen Zhang}, \bibinfo{person}{Yangang
  Wang}, \bibinfo{person}{Xiaoming Deng}, \bibinfo{person}{Yinda Zhang},
  \bibinfo{person}{Ping Tan}, \bibinfo{person}{Cuixia Ma}, {and}
  \bibinfo{person}{Hongan Wang}.} \bibinfo{year}{2021}\natexlab{}.
\newblock \showarticletitle{Interacting two-hand {3D} pose and shape
  reconstruction from single color image}. In
  \bibinfo{booktitle}{\emph{Proceedings of the IEEE International Conference on
  Computer Vision}}. \bibinfo{pages}{11354--11363}.
\newblock


\bibitem[Zhao et~al\mbox{.}(2023)]%
        {zhao2023best}
\bibfield{author}{\bibinfo{person}{Weichao Zhao}, \bibinfo{person}{Hezhen Hu},
  \bibinfo{person}{Wengang Zhou}, \bibinfo{person}{Jiaxin Shi}, {and}
  \bibinfo{person}{Houqiang Li}.} \bibinfo{year}{2023}\natexlab{}.
\newblock \showarticletitle{BEST: BERT Pre-Training for Sign Language
  Recognition with Coupling Tokenization}.
\newblock \bibinfo{journal}{\emph{Proceedings of the AAAI Conference on
  Artificial Intelligence}}, \bibinfo{pages}{3597--3605}.
\newblock


\bibitem[Zhou et~al\mbox{.}(2020)]%
        {zhou2020monocular}
\bibfield{author}{\bibinfo{person}{Yuxiao Zhou}, \bibinfo{person}{Marc
  Habermann}, \bibinfo{person}{Weipeng Xu}, \bibinfo{person}{Ikhsanul Habibie},
  \bibinfo{person}{Christian Theobalt}, {and} \bibinfo{person}{Feng Xu}.}
  \bibinfo{year}{2020}\natexlab{}.
\newblock \showarticletitle{Monocular real-time hand shape and motion capture
  using multi-modal data}. In \bibinfo{booktitle}{\emph{Proceedings of the IEEE
  Conference on Computer Vision and Pattern Recognition}}.
  \bibinfo{pages}{5346--5355}.
\newblock


\bibitem[Zimmermann and Brox(2017)]%
        {zimmermann2017learning}
\bibfield{author}{\bibinfo{person}{Christian Zimmermann} {and}
  \bibinfo{person}{Thomas Brox}.} \bibinfo{year}{2017}\natexlab{}.
\newblock \showarticletitle{Learning to estimate {3D} hand pose from single
  {RGB} images}. In \bibinfo{booktitle}{\emph{Proceedings of the IEEE
  International Conference on Computer Vision}}. \bibinfo{pages}{4903--4911}.
\newblock


\end{thebibliography}


\end{document}